\documentclass{article}

\usepackage{microtype}
\usepackage{graphicx}
\usepackage{subfigure}
\usepackage{booktabs} 
\usepackage{hyperref}
\usepackage{url}
\usepackage{bbm}
\usepackage{graphicx}    
\usepackage{subcaption}  
\usepackage[linesnumbered,ruled,vlined]{algorithm2e}

\usepackage{hyperref}

\usepackage[accepted]{icml2025}

\usepackage{amsmath}
\usepackage{amssymb}
\usepackage{mathtools}
\usepackage{amsthm}

\usepackage[capitalize,noabbrev]{cleveref}

\theoremstyle{plain}

\theoremstyle{definition}

\theoremstyle{remark}

\usepackage[textsize=tiny]{todonotes}

\icmltitlerunning{Discovering Chunks in Neural Embeddings for Interpretability}

\begin{document}

\twocolumn[


\icmltitle{Discovering Chunks in Neural Embeddings for Interpretability}



\icmlsetsymbol{equal}{*}

\begin{icmlauthorlist}
\icmlauthor{Shuchen Wu}{eml,mpi}
\icmlauthor{Stephan Alaniz}{eml,mcml,tum}
\icmlauthor{Eric Schulz}{hcai,mpi}
\icmlauthor{Zeynep Akata}{eml,mcml,tum}
\end{icmlauthorlist}

\icmlaffiliation{eml}{Explainable Machine Learning Lab, Helmholtz Munich}
\icmlaffiliation{tum}{Department of Computer Science, Technical University of Munich}
\icmlaffiliation{hcai}{Institute for Human-Centered AI, Helmholtz Munich}
\icmlaffiliation{mpi}{Max Planck Institute for Biological Cybernetics}
\icmlaffiliation{mcml}{Munich Center for Machine Learning}

\icmlcorrespondingauthor{Shuchen Wu}{shuchen.wu@tue.mpg.de}
\icmlkeywords{Interpretability, XAI, Cognitive Science, LLM}
\vskip 0.3in
]



\printAffiliationsAndNotice{}  

\begin{abstract}
Understanding neural networks is challenging due to their high-dimensional, interacting components. Inspired by human cognition, which processes complex sensory data by chunking it into recurring entities, we propose leveraging this principle to interpret artificial neural population activities. Biological and artificial intelligence share the challenge of learning from structured, naturalistic data, and we hypothesize that the cognitive mechanism of chunking can provide insights into artificial systems.
We first demonstrate this concept in recurrent neural networks (RNNs) trained on artificial sequences with imposed regularities, observing that their hidden states reflect these patterns, which can be extracted as a dictionary of chunks that influence network responses. Extending this to large language models (LLMs) like LLaMA, we identify similar recurring embedding states corresponding to concepts in the input, with perturbations to these states activating or inhibiting the associated concepts.
By exploring methods to extract dictionaries of identifiable chunks across neural embeddings of varying complexity, our findings introduce a new framework for interpreting neural networks, framing their population activity as structured reflections of the data they process.
\end{abstract}
\section{Introduction}


Neural networks are known as ``black box systems'' \cite{templeton-2021-word,petsiuk_rise_2018,ribeiro_why_2016-1,zhang_visual_2018,marks_sparse_2024,elhage2022toy}: as their computation are performed by millions and billions of interacting components - a number far exceeding any conventional models such as physical laws describing the interaction between bodies, or linear regression functions. Many approaches have attempted to bridge the gap between simple and complex models by replacing complex computations with symbolic substitutes. However, this is inherently difficult and inevitably involves a trade-off between computational efficiency and the number of symbolic components used \cite{adadiberrada,lipton2017mythosmodelinterpretability,belinkov-2022-probing}. 

\begin{figure}[t!]
\begin{center}
\centerline{\includegraphics[width=\columnwidth]{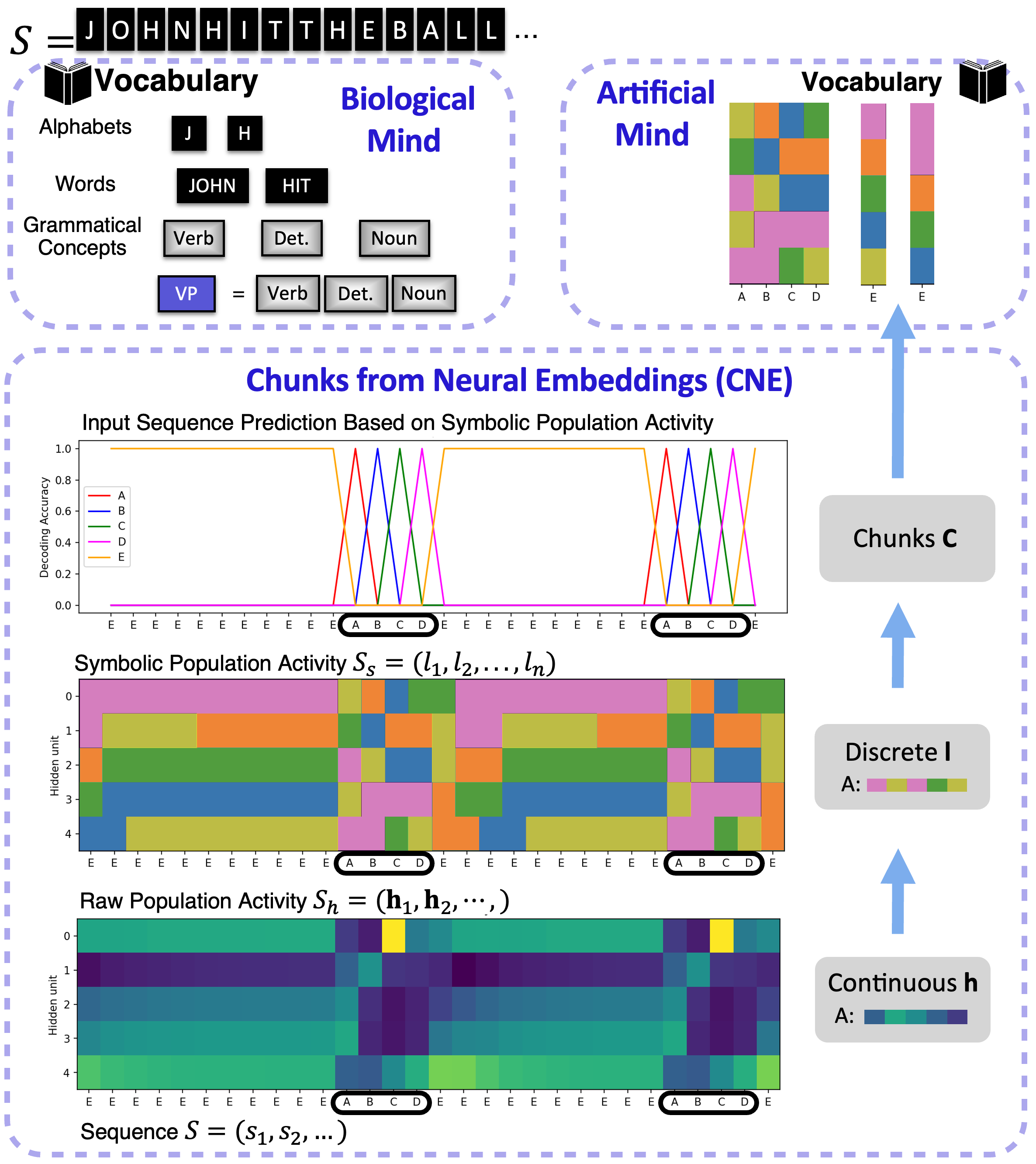}}
\caption{(Top) Naturalistic data is highly redundant and compositional, e.g. in language sequences. Cognitive systems segment redundancies by chunking recurring patterns. \textit{The reflection hypothesis} posits that ANNs neural activities can be interpreted as chunks that reflect the structured regularities in reality. (Bottom) In simple networks that contain a small number of neurons, chunking methods can be used to learn a dictionary of frequently recurring population trajectories. The discrete representations of the embedding state can reliably predict the input in the sequence and network's predictions.}
\label{fig:schematic}
\end{center}
\vskip -0.4in
\end{figure}


A substantial part of the interpretability question is cognitive \cite{adadiberrada,miller2018explanationartificialintelligenceinsights} and demands understanding what makes high dimensional data meaningful for people. We present a novel perspective on interpreting artificial neural networks by referencing how cognition interprets sensory data. Similar to the dimensionality of neural networks' activities, perceptual data that flood into our sensory stream is also high-dimensional. We make sense of this ``booming, buzzing confusion'' by instantly segmenting recurring patterns in perceptual sequences as chunks \cite{Graybiel1998TheRepertoires, Gobet2001ChunkingLearning, Egan1979ChunkingDrawings, Ellis1996SequencingOrder, koch_patterns_2000, Chase&Simon1973}. In this way, cognition perceives a structured representation of the overwhelming flow of the sensory stream as the appearance and disappearance of chunks over time \cite{wu_chunking_2023,wu2025motifs}. Inspired by this principle of cognition, we explore whether the cognitive mechanism of chunking can be leveraged to interpret high-dimensional neural population activities.

We hypothesize that this approach could work if the neural network’s hidden activity reflects the regularities in the data it learns to predict.  To test this ``reflection hypothesis'', we first engineer datasets by injecting known regularities and analyze the neural population activity of a simple RNN trained to predict these sequences as a proof of concept. We then extend our investigation to large language models predicting sequences. Throughout, we explore several chunk extraction methods from low to high dimensions of neural population activities, highlighting their strengths and limitations. Our findings suggest a novel interpretability framework that leverages the cognitive principle of chunking to identify meaningful entities within artificial neural activities.



\section{Related Work}
Most interpretability approaches hold a salient agreement on ``what is interpretable'' and ``what to interpret''. ``What is interpretable'' is influenced by models in physics and mathematics, where operations and derivations are framed around the manipulation of a small set of well-defined symbols. Hence, interpretable concepts are confined to word-level or token-level description, and approaches try to learn a mapping between the neural activities and the target interpretable concept to understand \cite{geva_transformer_2022, zou_representation_2023, belinkov-2022-probing,belrose2023tunedlens,pal2023futurelens,yomdin2023jump}. 

The current approaches on ``what to interpret'' to understand the computations inside a neural network is heavily influenced by neuroscience: either on the level of neurons as a computation unit or in a low-dimensional neural activity descriptions. The earliest interpretability approaches, inspired by neuroscience discoveries such as ``grandmother cells'' and ``Jennifer Aniston neurons'', focused on understanding the semantic meanings that drive the activity of individual neurons. Similarly, studies in artificial neural networks, from BERT to GPT, have identified specific neurons and attention heads whose activations correlate with semantic meanings in the data \cite{olah2020neuron, elhage2022toy, wang2022interpretabilitywildcircuitindirect, marks2024sparsefeaturecircuitsdiscovering, bau2020concept, goh2021multimodal, nguyen2016synthesizing, mu2021compositionalexplanationsneurons, radford2017learninggeneratereviewsdiscovering}. The sparse autoencoders (SAEs) approach can be seen as an intermediate step that encourage the hidden neurons to be more monosemantic \cite{bricken2023interpretable,braun2024identifyingfunctionallyimportantfeatures, cunningham2023sparseautoencodershighlyinterpretable,chaudhary2024evaluatingopensourcesparseautoencoders,karvonen2024evaluatingsparseautoencoderstargeted}. Thereby, one trains an autoencoder to map neural activities of a hidden unit layer to a much larger number of intermediate hidden units while encouraging a sparse number of them to be active. In this way, the target hidden layer activity can be represented by a superposition of several individual neurons inside the SAE. 
Other approaches reduces and interprets neural population activities in lower dimensions: representation engineering  captures the distinct neural activity corresponding to the target concept or function, such as bias or truthfulness \cite{zou_representation_2023}. Then, it uses a linear model to identify the neural activity direction that predicts the concept under question or for interference with the network behavior. 

The current interpretability approach that studies language-based descriptions as conceptual entities and their implications for individual/low-dimensional neurons suffers from limitations on both ends: meanings are finite, and individual neurons are limited in their expressiveness and may not map nicely to these predefined conceptual meanings. 

Just like physics models lose their failure to have a closed-form description of motion beyond two interacting bodies \cite{tao_e_2012}, confined, symbolic definitions of interpretation have inherent limitations in precision. This cognitive constraint—our reliance on well-defined symbolic entities for understanding—has made deciphering the complexity of billions of neural activities an especially daunting task. It underscores a fundamental trade-off between the expressiveness of a model and its interpretability \cite{wang2024largelanguagemodelsinterpretable}.

Focusing solely on individual neurons is also is insufficient to capture the broader mechanisms underlying neural activity across a network. ``monosemantic'' neurons, which respond to a single concept, make up only a small fraction of the overall neural population \cite{radford2017learninggeneratereviewsdiscovering, elhage2022toy, wang2022interpretabilitywildcircuitindirect, dai2022knowledge, voita2023neuron,miller2023neuron}. Empirically, especially for transformer models \cite{elhage2022toy}, neurons are often observed to be ``polysemantic'',, i.e., associated with multiple, unrelated concepts \cite{mu2021compositionalexplanationsneurons, elhage2022toy, olah2020circuits}, which complicates the task of understanding how neural population activity evolves across layers \cite{elhage2022toy, gurnee2023findingneuronshaystackcase}. This highlights the need for more holistic approaches that account for the complex, distributed nature of neural representations.

\section{Chunks from Neural Embeddings (CNE)}
We effortlessly perceive high-dimensional perceptual signals by segmenting them into recurring, meaningful patterns. The recurring patterns reside in a subpopulation of our perceptual dimensions as cohesive wholes \cite{Miller1956TheInformation, laird1984towards,Graybiel1998TheRepertoires,Gobet2001ChunkingLearning}.


Humans segment perceptual data into chunks as a strategy to reduce the complexity of naturalistic data.
As naturalistic data exerts a compositional structure and contains rich regularities - isolating out the recurring entity as a concept, is an effective way for an agent to compress their observation into entities and their relations \cite{wu_chunking_2023, wu_learning_2022}. For instance, as shown in Figure \ref{fig:schematic} (top-left), natural language contains recurring alphabets, words, phrases, grammatical rules, and morphological and sentence structures on a concrete and abstract level. 

Similar to humans, artificial intelligence systems are tasked with this problem of understanding reality from naturalistic data. The regularities in naturalistic data may drive converging representations in diverse AI models. Past research indicates that neural networks of different architectures, scales, and sizes learn representations that are remarkably similar \cite{balestriero18b, moschella2023relativerepresentationsenablezeroshot}, even when trained on different data sources \cite{lenc2015understandingimagerepresentationsmeasuring}. This phenomenon is particularly pronounced in larger, more robust models\cite{bansal21,dravid2023rosettaneuronsminingcommon,kornblith19a,roeder21a}. Such patterns have been observed across multiple data modalities \cite{lenc2015understandingimagerepresentationsmeasuring,moschella2023relativerepresentationsenablezeroshot,kornblith19a,roeder21a}. Recently, it has been hypothesized that such convergences reflect an underlying statistical model of reality, aligning with a Platonic conceptual framework \cite{huh2024platonicrepresentationhypothesis}.

\textbf{The Reflection Hypothesis}:
We propose that artificial neural networks, like human cognition, may reflect the regularities and redundancies present in the data they are trained on by encoding these patterns within their neural computations. Specifically, we hypothesize that a well-trained neural network should exhibit trajectories of neural computation that mirror the structure of its training data.

If this hypothesis holds, it could allow us to leverage the human cognitive ability to ``chunk'' high-dimensional perceptual data into meaningful units. Just as human perception breaks down complex sensory inputs into recurring, segregated patterns, neural network activities might also be decomposable into recurring "chunks". These chunks, residing within subsets of all neurons of the network, could represent distinct patterns of computation and serve as interpretable entities. By identifying and analyzing these chunks, we could reduce the complexity of neural activities into interactions between these interpretable units.

To test this hypothesis, we develop and explore methods to extract these chunks across a spectrum of settings, starting from simple models trained on straightforward sequences to complex models trained on intricate, real-world data. This approach aims to bridge the gap between human cognition and the interpretability of artificial neural networks.

\subsection{Extracting Invariances}
We explore multiple methods to extract recurring chunks in neural population activities, which can be applied to different dimensionalities of data.
Formally, denote the training sequence as $S = (s_1, s_2, \dots, s_{n})$, indexed by $I = \{1,2,3,\cdots, n\}$, the sequence of neural network activations of the input sequence as 
$S_h = (\mathbf{h}_1,\mathbf{h}_2,\cdots, \mathbf{h}_n )$, which we also refer to as neural population activity.
Each neural population vector has embedding dimension $d$: $\mathbf{h}_i \in \mathbb{R}^d$. 
Depending on the dimensionality $d$ and the nature of the problem, we develop three chunk extraction methods: \textbf{discrete sequence chunking} for small $d$, \textbf{neural population averaging} for large $d$ and when there is an identifiable pattern in $S$, and \textbf{unsupervised chunk discovery} when no supervising pattern is available.
\begin{figure}[t]
\begin{center}
\centerline{\includegraphics[width=\columnwidth]{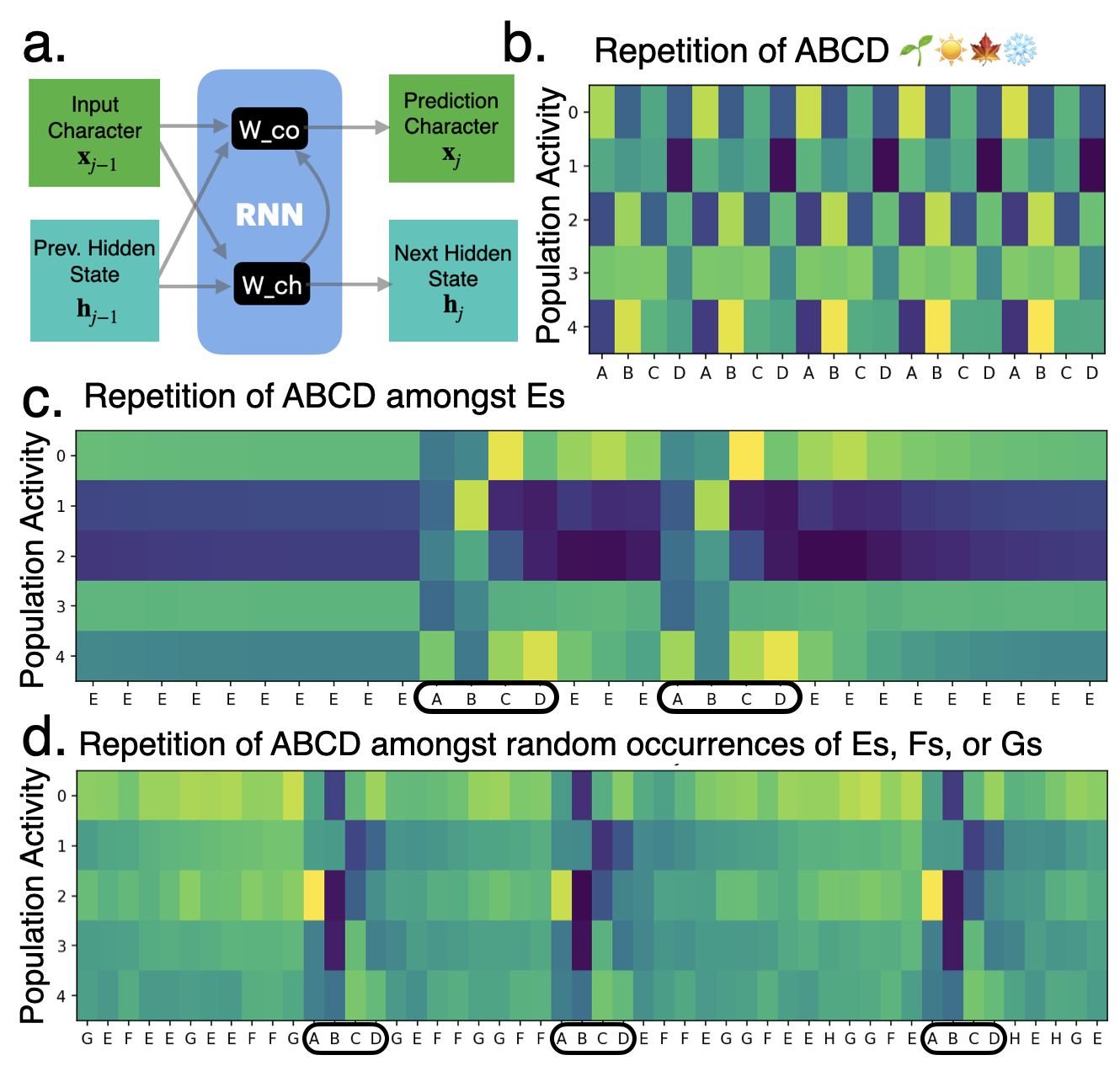}}
\caption{Testing the reflection hypothesis with simple RNNs and artificial sequences.
a. RNN updates predictions and memory states based on inputs and previous hidden state.
b. Neural population activity (of the first 5 neurons) in response to repeating chunk (ABCD); c. Sparse occurrence of ABCD within a default sequence (E);
d. ABCD persists as a cohesive chunk amid background noise (random E, F, G). }
\label{fig:RNN}
\end{center}
\vskip -0.4in
\end{figure}
\paragraph{Discrete Sequence Chunking}
When $d$ is relatively small, we can use a cognitive-inspired method to extract recurring patterns \cite{wu_learning_2022}, which is also used for text compression \cite{gage1994new,zaki2000sequence,agrawal1995mining} or  tokenization. The idea is to cluster the neuron activations individually, use the cluster indices to represent the network activity as strings, and then further group frequently subsequently occurring strings into chunks. For this purpose, we convert each hidden neural activity vector $\mathbf{h} \in \mathbb{R}^d$ into a string of discrete integers (example shown in Figure \ref{fig:schematic} (bottom). In this way, the sequence of vectors $S_h$ is transformed into a one-dimensional sequence of strings $S_s = (l_1, l_2, ..., l_n) $. Each $l_j$ has length $d$ and contains the nearest cluster index for each of the $d$ neurons, i.e., $
l_j = \text{concat}(\mathcal{C}(h_1), \mathcal{C}(h_2), \dots, \mathcal{C}(h_d))$. $\mathcal{C}$ is a clustering function that assigns the closest cluster index to the activity of the neuron.  

We then apply a chunking method on $S_s$ to extract a vocabulary of chunks manifested in patterns of string combinations, representing patterns of neural state trajectories. The vocabulary of chunks $\mathbf{D}$ is initialized with the set of unique strings in $S_s$.
The vocabulary expands with newly learned chunks for a number of iterations. In each iteration, the $n$ most frequently occurring chunk pairs above a frequency threshold are concatenated and added to the vocabulary. The updated vocabulary is then used to parse the sequence in the next iteration for another round of chunk merging until convergence (pseudocode is described in appendix \ref{alg:Learnchunks}). By merging patterns that occur often adjacently, we identify the frequently recurring population trajectories. 


\begin{figure}[t]
\centering
\begin{subfigure}
  \centering
  \includegraphics[width=0.49\columnwidth]{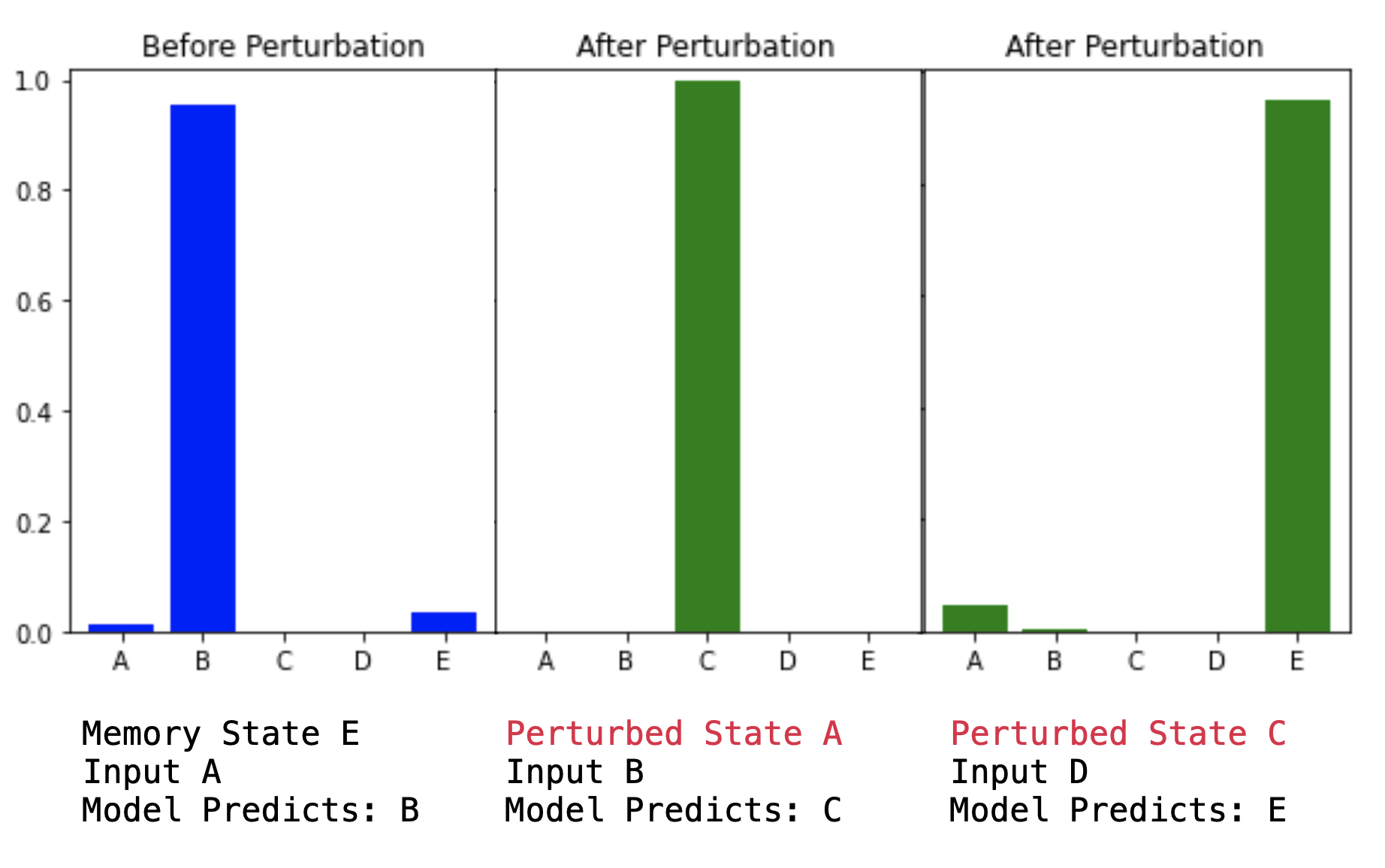}
\end{subfigure}%
\begin{subfigure}
  \centering
  \includegraphics[width=0.49\columnwidth]{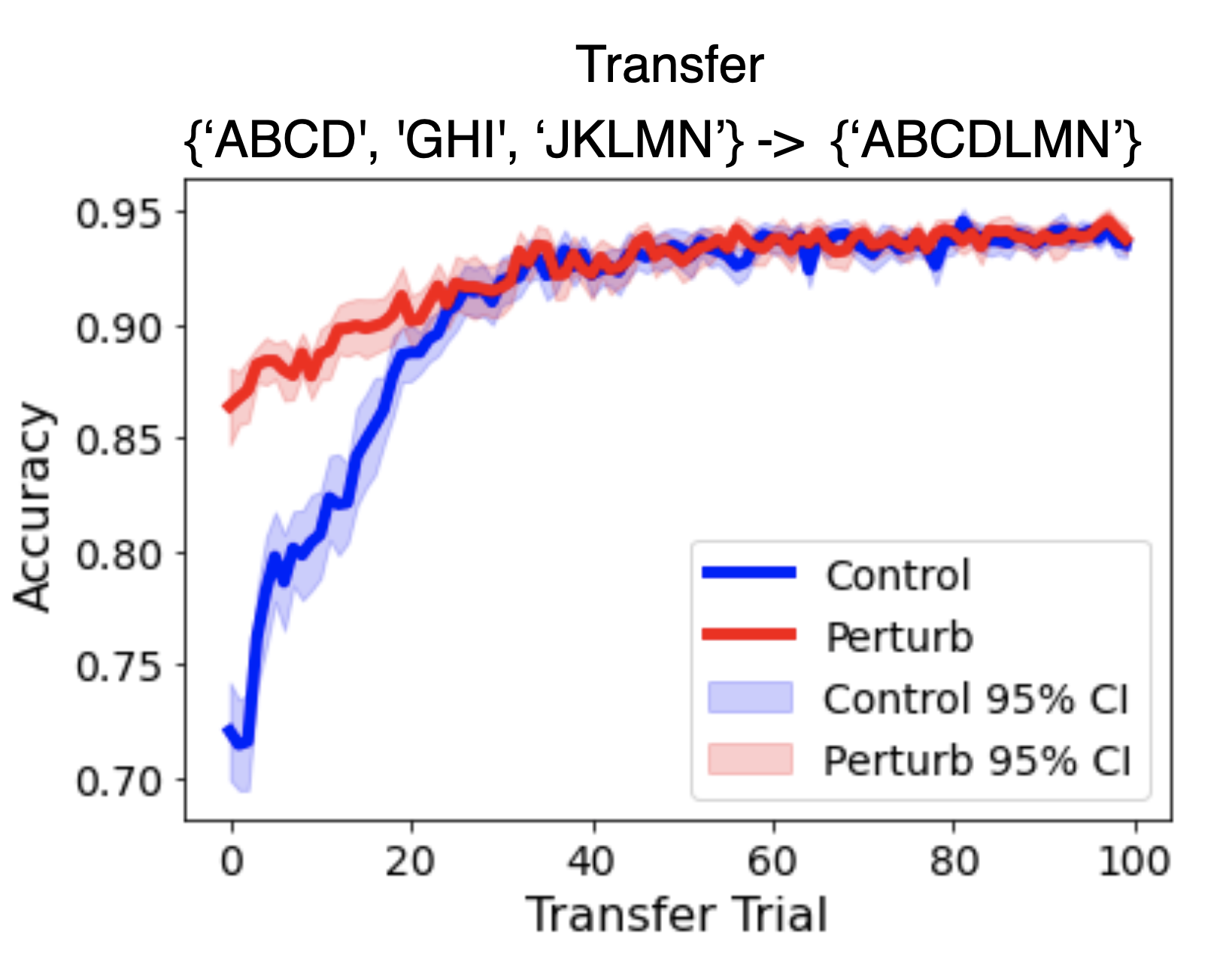}
\end{subfigure}
\caption{Left: Hidden states can be grafted to causally change network memory and prediction. Right: Embedding grafting enables faster transfer learning of a compositional vocabulary.}
\label{fig:rnnperturb_a}
\end{figure}

\begin{figure}[t]
\centering
\begin{subfigure}
  \centering
  \includegraphics[width=0.49\columnwidth]{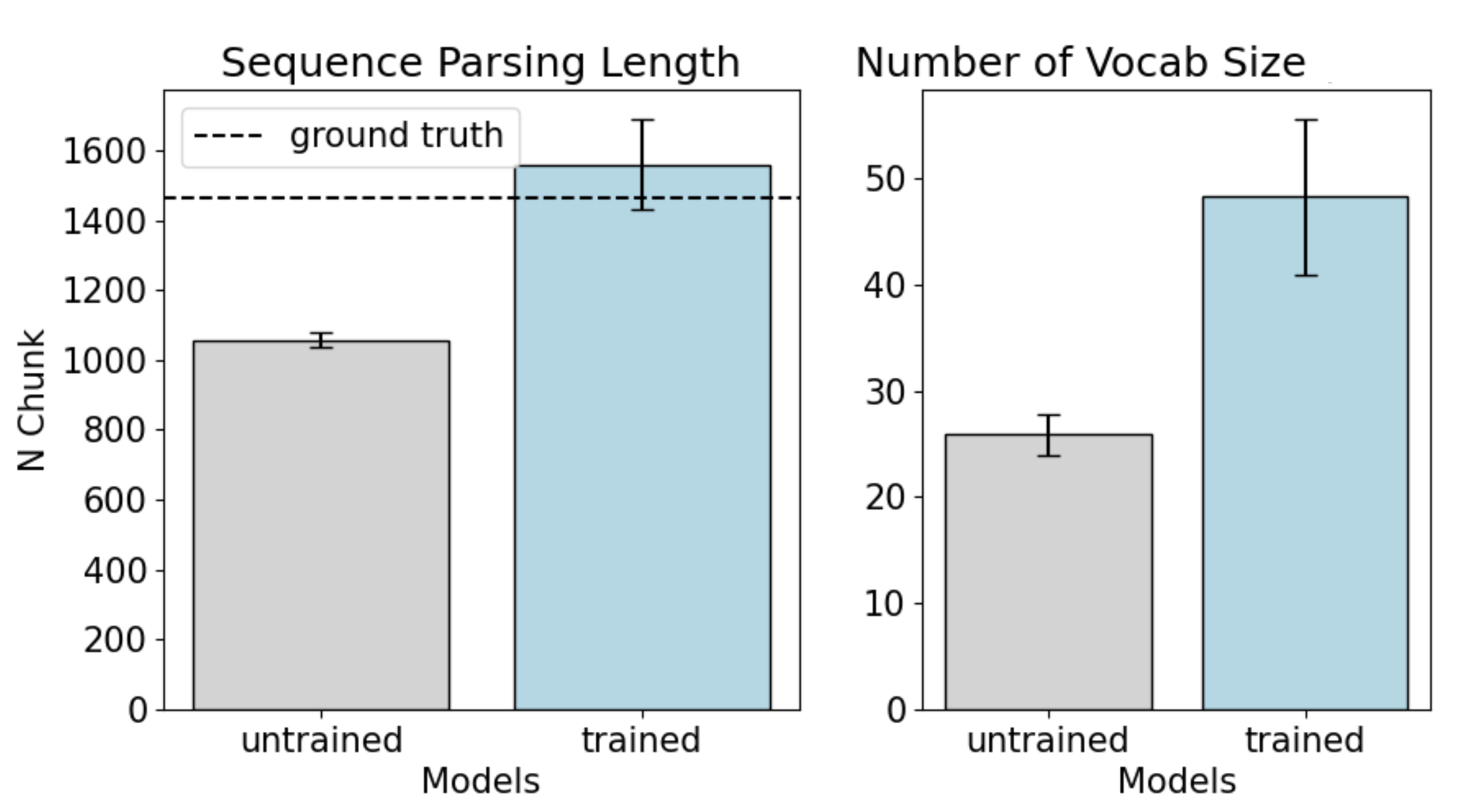}
\end{subfigure}%
\begin{subfigure}
  \centering
  \includegraphics[width=0.49\columnwidth]{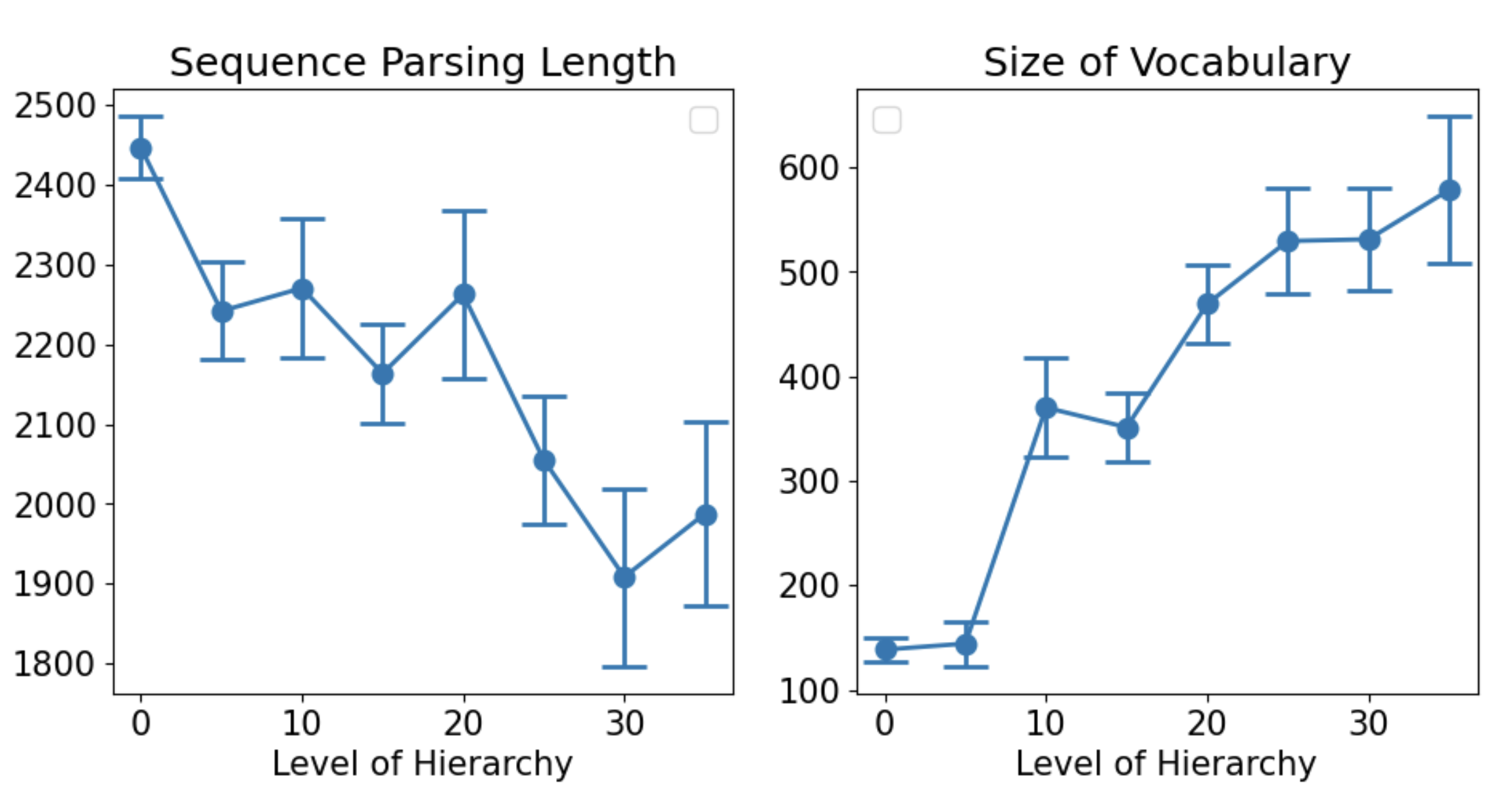}
\end{subfigure}
\caption{Left: Training creates extra chunks inside the embedding space. Right: The number of embedding states increases with the complexity of the input sequence.}
\label{fig:rnnperturb_b}
\end{figure}


\paragraph{Neural Population Averaging}
In larger-scale networks like transformers, $d$ can be very big. Moreover, complex sequences — such as natural language — exhibit patterns at both surface and more abstract levels, such as words or category of word-types. In this scenario, identifying recurring patterns in the sequence allows us to uncover the internal structures of neural population chunks by averaging the relevant neural population activities, akin to extracting task-induced neural response functions \cite{Cohen2011,Engel2001,Churchland2012}.

We assume that there are also chunks, i.e., recurring activities of neural subpopulations, which account for the network's computation as elicited by a recurring pattern $s$ in the sequence (such as a particular word `cheese'). 
Denote the indices of the pattern encoding chunks as $C(s) \subseteq W$, i.e., a subset of the neurons in $\mathbf{h}$ among the whole neuron population $W$ ($|W| = d$). Denote the set of indices where the pattern appears in the input sequence as $V(s) \subset I$. From the Strong Law of Large Numbers (SLLN), the population mean of signal-encoding neural subpopulation activities $\mathbf{h}_{C(s)}$ converges to the true mean as $|V(s)|$ approaches infinity:
\begin{equation}
    \overline{\mathbf{h}_{C(s)}} = \frac{\sum_{j \in V(s)} \mathbf{h}_{C(s),j}}{|V(s)|} \text{and} \lim_{|V(s)| \to \infty} \overline{\mathbf{h}_{C(s)}} = \boldsymbol{\mu} \quad \text{(a.s.)}
\end{equation}
where $\boldsymbol{\mu} = \mathbb{E}[\mathbf{h}_{C(s)}]$ is the true mean of the subpopulation neural activity.

Given training data containing the sequence \( S \), the corresponding neural population activity \( S_h \), and a recurring pattern \( s \) in \( S \), we aim to estimate the signal-relevant chunk. This includes the mean neural subpopulation activity \( \overline{\mathbf{h}_{C(s)}} \) within the subpopulation neurons \( C(s) \subset W \), as well as the range of deviation $\Delta$ within which the neural activities \( \mathbf{h}_{C(s)} \) fluctuate around \( \overline{\mathbf{h}_{C(s)}} \).

To do this, we first compute the mean population response of signal $s$ by averaging over the token-specific hidden state representations $\overline{\mathbf{h}} = \frac{\sum_{j \in V(s)} \mathbf{h}_{(j)}}{|V(s)|}$. A neuron $i$ is hypothesized to be inside ${C(s)}$ if its activity at the time of the pattern-specific index fluctuates within a pre-set tolerance level around the mean signal-relevant activity $C(s) = \{ i \in W : |h_{i,j} - \overline{h_i}| \leq \text{tol} \quad \forall j \in V(s)\}$.

After estimating the neural subpopulation $C(s)$ and $\overline{\mathbf{h}_{C(s)}}$, we then calculate a maximal deviation acceptable in the training data:
\begin{equation}
    \Delta = \max_{j \in V(s)} \frac{||\mathbf{h}_{{C(s),j}} - \overline{\mathbf{h}_{C(s)}}||_2^2}{d}
\end{equation}
We define the deviation \( \Delta \) as a threshold for acceptable fluctuation around the mean neural subpopulation activity \( \overline{\mathbf{h}_{C(s)}} \). An unknown neural activity vector \( \mathbf{h} \) is classified as belonging to the chunk \( \overline{\mathbf{h}_{C(s)}} \) if it lies within the closed ball $\overline{B}$ centered at \( \overline{\mathbf{h}_{C(s)}} \) with radius \( \Delta \), i.e.,  

\begin{equation}
\mathbf{h} \in \overline{B}(\overline{\mathbf{h}_{C(s)}}, \Delta) \quad \text{iff} \quad \frac{\|\mathbf{h}_{C(s)} - \overline{\mathbf{h}_{C(s)}}\|_2^2}{d} \leq \Delta.
\end{equation}
Using this criterion, we identify instances of an unknown neural subpopulation vector $\mathbf{h}_{C(s)}$ in the test data using the chunk identification function
\begin{equation}
f_{\text{chunk}}(\mathbf{h}_{C(s)}, \overline{\mathbf{h}_{C(s)}}, \Delta(s)) =
\begin{cases} 
1, & \text{if } \frac{\|\mathbf{h}_{C(s)} - \overline{\mathbf{h}_{C(s)}}\|_2^2}{d} \leq \Delta(s), \\ 
0, & \text{otherwise}.
\end{cases}
\end{equation}
and evaluate its quality as a neural population activity classifier for the occurrence of the signal \( s \). This evaluation is based on the true positive rate (TPR) and false positive rate (FPR).

Since the neural subpopulation \( C(s) \), the chunk \( \overline{\mathbf{h}_{C(s)}} \), and the deviation threshold \( \Delta \) depend on a tolerance parameter, we generate a series of increasingly stringent tolerance thresholds: $\text{tol}_i = 2 \times 0.8^i, \quad i = 0, 1, \dots, 39,$
and optimize the tolerance parameter to maximize TPR while minimizing FPR on the training data.


This method relies on prior knowledge of recurring patterns $s$ in the input sequence and assumes that the network's activity remains within a small deviation threshold around the mean chunk activity vector, $\overline{\mathbf{h}_{C(s)}}$. If the network generates multiple chunks to distinguish contextual variations, the signal's identifiability may decrease due to the averaging, potentially limiting its utility in more complex or highly dynamic contexts.

\paragraph{Unsupervised Chunk Discovery}
\label{med:unsupervised}
While discrete sequence chunking is limited to a small dimensionality $d$ and population averaging hinges on the knowledge of existing patterns in the input sequence, we also try learning chunks in high dimensional embedding space without the knowledge of the recurring signal in the input. 

We formulate the chunk finding question as finding clusters in the embedding space by learning an embedding dictionary \( \mathbf{D} \in \mathbb{R}^{K \times d} \) which contains $K$ chunks, each with embedding dimension $d$.

From the embedding data \( \mathbf{X} \in \mathbb{R}^{M \times d} \) of a layer from an LLM processing some tokenized input with batch size $M$ and embedding dimension $d$, we want to find chunks as recurring entities in the embedding data. To do this, we optimize a loss function \(\mathcal{L}(\mathbf{X},\mathbf{D})\) encouraging the similarity between the embedding data and the maximally similar chunk in the dictionary:
\begin{equation}
    \mathcal{L} = - \frac{1}{M} \sum_{m \in \{1, \cdots M\}}\max_{k \in \{1, ..., K\}}\mathbf{SIM}(\mathbf{D}_k, \mathbf{X}_m)
\end{equation}
We compute the pairwise normalized cosine similarity between embedding dictionary entries and embedding data: 
\[
\mathbf{SIM}(\mathbf{d}, \mathbf{x}) = \frac{\mathbf{d}^\top \mathbf{x}}{\|\mathbf{d}\|_2 \|\mathbf{x}\|_2},
\]
where \( \|\mathbf{d}\|_2 \) is the \( L_2 \)-norm of \( \mathbf{d} \) and \( \|\mathbf{x}\|_2 \) is the \( L_2 \)-norm of \( \mathbf{x} \).





\section{Results}
\subsection{Proof of concept in simple RNNs}

First, we assessed the reflection hypothesis using a simple recurrent neural network (Figure \ref{fig:RNN}a, details in \cref{appendix:RNN}) trained on artificially generated sequences with a known chunk, ABCD. We begin with sequences featuring periodic occurrences of ABCD (Figure \ref{fig:RNN}b), then gradually increase complexity and noise: first, by embedding ABCD sparsely within a default sequence of E (Figure \ref{fig:RNN}c), and then, by presenting ABCD as a cohesive chunk amid background noise composed of random occurrences of E, F, and G (Figure \ref{fig:RNN}d). For each sequence type, an RNN was trained to predict the next element in the sequence. Visually, the hidden states of the RNN exhibited identifiably recurring patterns aligned with the ABCD chunks in the input sequence. 

\paragraph{Understanding the neural state and its mapping to the input}
We then applied \textit{Discrete Sequence Chunking} (\cref{fig:schematic} bottom) to convert hidden unit activity from a continuous time scale to the symbolic and discretized description. Thereby, we can denote population activities by the alternation of indices that marks the belonging cluster centroid and visualize the population activity by the indices of the assigned nearest cluster. 
This symbolic description of the neural population trajectory allows decoding the trajectory of the hidden state and its corresponding input regularity via a look-up table (illustrated in \cref{appendix:lookuptable}), and reaches a perfect decoding accuracy on the test set. 

From this the symbolic description of neural population trajectories, we can then apply chunk discovery methods to learn a dictionary of symbolized chunks. This dictionary contains the maximally recurring patterns inside the network, reflecting the patterns in the data consistently and in an interpretable way. This suggests that RNN's hidden unit activities when trained on sequences with repeating inputs, follow a highly regularized trajectory, and conventional chunking methods can extract the neural activity corresponding to processing these sequence regularities. 

\paragraph{Population activity chunks causally alter network's behavior}
We test the causal role of these extracted recurring population states by grafting the embedding state to the identified chunk states. 
As shown in Figure \ref{fig:rnnperturb_a} (left), we start with the current state of the RNN that predicts B from the input character A, and the memory state that resulted from the previous input E. Perturbing the hidden population to A, or C, while pairing the hidden population state with the input B, or D, will separately change the network's output to be C, or E, respectively. Thus, grafting the hidden population state to the identifiable regularity of the network's hidden state can lead to predictable behavioral changes. 
\begin{figure}[h]
\begin{center}
\centerline{\includegraphics[width=\columnwidth]{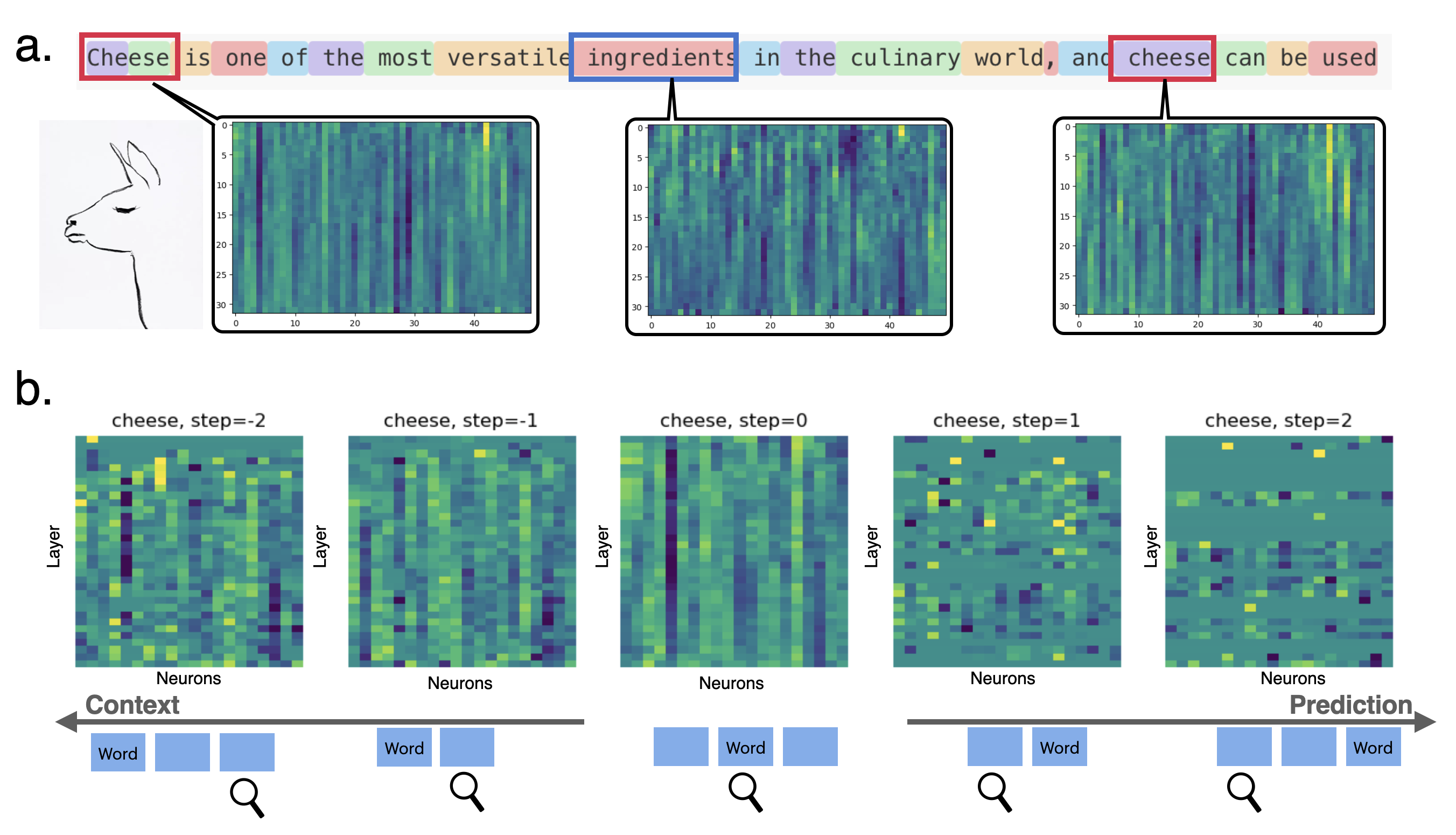}}
\caption{a. Neural embedding activity of the first 50 neurons (unselected) across all layers (33) processing prompt up until the end of each highlighted word. 
b. Extracted neural activity chunks in response to word at different sequence positions.}
\label{fig:llamademo}
\end{center}
\vskip -0.2in
\end{figure}
\paragraph{Grafting neural population activity in chunks facilitates compositional learning}
We test whether the population grafting may artificially induce the network to compose and reuse its previously learned vocabulary to learn a new vocabulary. We trained two identical RNNs on a synthetic training sequences with randomly sampled words from a dictionary \{ABCD, GHI, JKLMN\} occurring within null character sequences of Es. 

We then further train the RNNs in a transfer sequence with the word ABCDLMN - a composite of two-word parts in the training vocabulary. For one RNN, we construct a lookup table that maps the discrete population state to the concurrent input. 
As this RNN learns on the transfer sequence, whenever the input is D (the network state should be C), we graft the neural population state to the cluster centroid of the hidden state induced by the previous input being J paired with the input character being K, thereby forcing the network to generate the next prediction as L instead of E. Shown in Figure \ref{fig:rnnperturb_a} (right) are the learning curves between the two networks. The perturbed RNN has a better transfer sequence than the RNN without perturbation. Thus, grafting hidden states artificially can force the network to reuse its previously learned words to acquire a composite word. 
\paragraph{Training creates additional population states that distinguishes input patterns}
We further hypothesized that learning may create additional chunks of neural population activities to distinguish different contexts. Taking one example: when the sequence contains words sampled from dictionary: \{CDAB, AB, ABCD\} among a default sequence of Es. The AB standing alone contains a different meaning than the AB inside ABCD. While a network that is not trained may not distinguish between the two types of ABs, a well-trained network should distinguish between the two. As in the former case the network shall predict E, in the later case the network shall predict CD following AB. As shown in Figure \ref{fig:rnnperturb_b} (left), words in the input sequence induces similar number of chunks in RNN's neural state trajectory, which is higher than an untrained network, suggesting that the untrained network contains neural trajectory states that do not distinguish contexts. The trained network also processes neural trajectories that contain a more diverse set of chunks. This suggests that the RNN creates new neural states to distinguish contexts that consist of identical sequential parts that lead to different predictions for the upcoming sequence.  

\begin{figure}[ht]
\vskip -0.1in
\begin{center}
\centerline{\includegraphics[width=0.9\columnwidth]{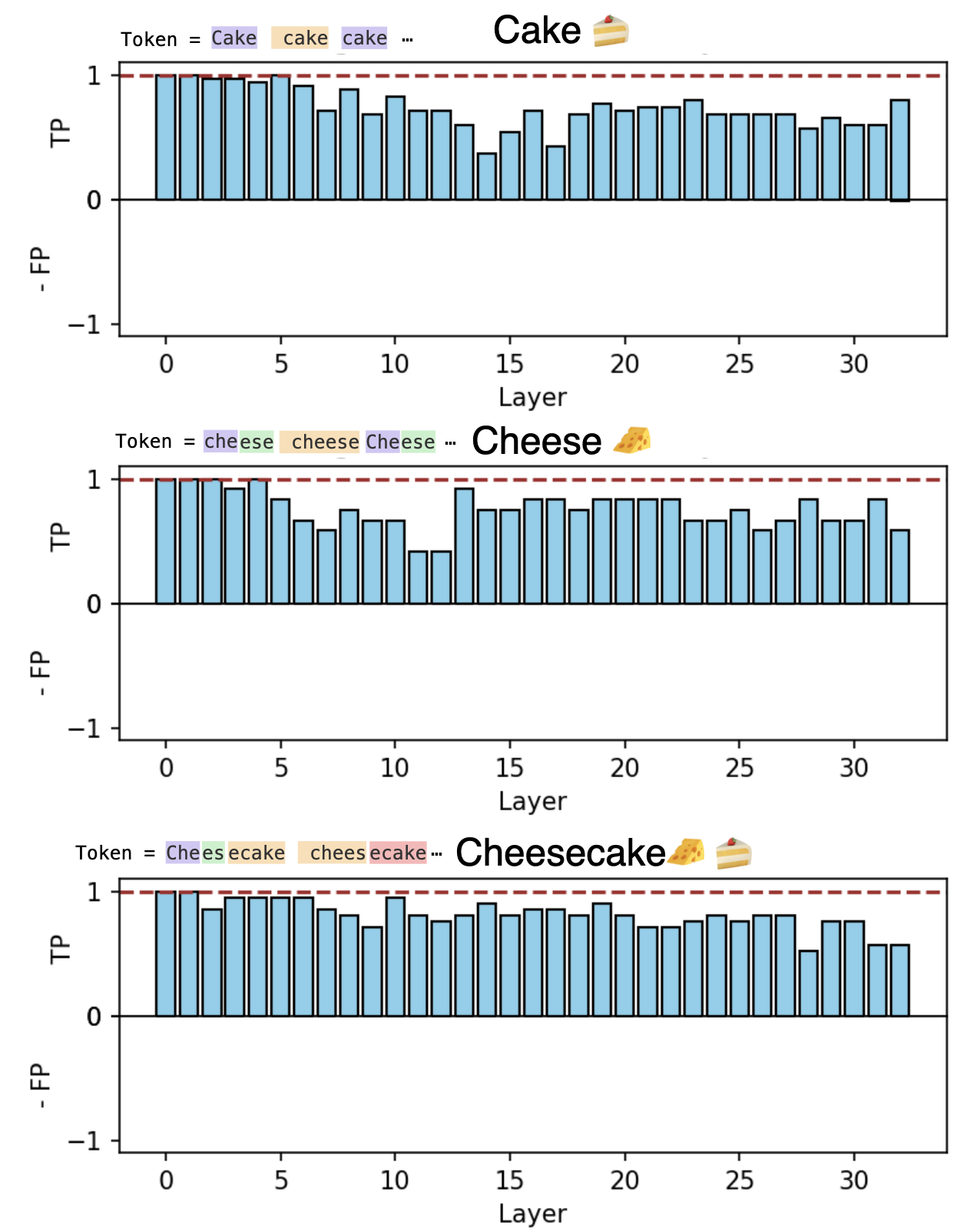}}
\caption{The identifiability of the presence of extracted chunks evaluated by signal detection measures.}
\label{fig:llamapopavg}
\end{center}
\vskip -0.4in
\end{figure}

\paragraph{Hierarchically structured input sequence corresponds to hierarchically structured states in neural state space}
Another implication of the reflection hypothesis is that when the input sequence contains an inherent hierarchical structure, i.e., alphabets can become word parts, which can become words, etc., then a neural network which learns to predict the sequence should also exhibit a more diverse set of population state manifested in more chunks. 

We generated a synthetic sequence with a hierarchical vocabulary as in \cite{wu_learning_2022} to test this prediction. The vocabulary is initialized with an alphabet set \{A,B,C,D\}, and E being the default null character. We then expands the vocabulary by generating new words in $n$ number of iterations. In each iteration, two words are randomly chosen from the existing vocabulary and concatenated to form a new word. A random probability vector is sampled from the flat Dirichlet distribution sorted in ascending order and scaled by 0.2 with E assigned a probability of 0.8. The final probabilities vector represents the likelihood of selecting each character from the vocabulary. The training sequence is generated by repeatedly sampling words from the vocabulary.

As shown in Figure \ref{fig:rnnperturb_b} (right), as the level of hierarchy increases, the vocabulary size expands, and so do the number of learned neural population chunks of the RNNs trained on such sequences. Meanwhile, the sequence concurrent neural population trajectory can be parsed by bigger and bigger neural trajectory chunks, suggesting that the nested hierarchies in the input sequence lead to more chunks in the neural state trajectories accounting for the hierarchically structured words in the sequence.  
\subsection{Reflection Hypothesis on LLMs}
As we have identified chunks in neural trajectories that influence small neural network's computation, we aim to further investigate the reflection hypothesis on larger neural networks. Specifically, we seek to determine whether any of our chunking methods enable an interpretable segmentation of an LLM’s neural population activities. To test this, we applied the reflection hypothesis to larger models, specifically LLaMA3-8B \cite{dubey2024llama}, while processing tokenized natural language sequences $S = (s_1, s_2, ..., s_n)$. We analyzed LLaMA3’s hidden states as it incrementally processed sequences—starting from the beginning of each sentence up to the current token. Given a prompt, the transformer predicts the next token based on the preceding context. By examining the hidden representations of the predicted token, we can track how the model's internal activity evolves as new tokens are added.
To capture this evolution, we recorded the hidden state sequence across all layers of LLaMA3 while it processed each token in the prompt, yielding a trajectory of neural population activities: $S_h =(\mathbf{h}_1, \mathbf{h}_2, ..., \mathbf{h}_n)$. We then examine the existence of neural chunks applying the \textit{Neural Population Averaging} and \textit{Unsupervised Chunk Discovery} methods for high dimensional neural data. 

\paragraph{Recurring words in a sentence}
Taking the word ``cheese'' as an example, Figure \ref{fig:llamademo}a visualizes the normalized hidden unit activities of the first 50 neurons across all LLaMA layers. Notably, network activity was more consistent with each other when ``cheese'' appeared as the last token than when a different word (e.g., ``ingredient'') was last. Despite incremental autoregressive processing and tokenization splits, both occurrences elicited similar states across all LLaMA layers, indicating the existence of neural subpopulation chunks responsible for the computation involving ``cheese'' in the prompt. 

Using recurring words in language as the pattern of interest \( s \), we applied \textit{the population averaging method} to identify the neural subpopulation \( C(s) \), the mean chunk activity \( \overline{\mathbf{h}_{C(s)}} \), and the fluctuation radius \( \Delta(s) \). For multi-token words, we defined the signal’s occurrence at the final token’s position. We assessed the extracted chunks and thresholds by measuring how well chunk detection predicted the occurrence of \( s \). Figure \ref{fig:llamapopavg} shows that the population averaging method yielded neural subpopulation chunks that are predictive of the word ``cheese'', ``cake'', and ``cheesecake'' in each layer and distinct from another, with low false positives and high true positives. Earlier layers were more faithful than later ones, likely because later layers introduce additional processing to distinguish contextual variations.


We can then generalize this method to other locations of the token, such as prior to or following the occurrences of the signal at indices $V(s)$, denote the set of indices undergoing this $k$ step shift as $V^{-k}(s) = \{ t - k \mid t \in V(s) \}, \quad V^{+k}(s) = \{ t + k \mid t \in V(s) \}$. The former ($k>0$) examines the network's neural activity representing a memory of the signal, and the latter ($k<0$) examines the network's neural activity that is predictive of signals happening subsequent to the input prompt sequence. 

Figure \ref{fig:llamademo}b visualizes the subpopulation activity chunks extracted by the population-averaging method responding to the words ``cheese'' at the current, previous, and subsequent signal indices. One observation is that the information about context is represented in the later rather than the earlier layer of the network. Meanwhile, information about the memory of a signal was represented much more sparsely among the neurons in the network than the most recent signal. Additionally, the temporal coding of the signal is not uniform, representing the same word `cheese' as memory, as the latest token, or as prediction is manifested in distinct neural population activities. There is a smaller neural population responsible for predicting a future signal than encoding for the current signal or the past signal. 

Generalizing this method to other words, we extracted neural subpopulation activity chunks accounting for the 100 most frequently occurring words in English, ranging the context and prediction step up to two more tokens. Appendix \ref{appendix:top100} shows more examples of evaluations on chunks extracted by this method.

\begin{figure}[ht]
\vskip -0.1in
\begin{center}
\centerline{\includegraphics[width=\columnwidth]{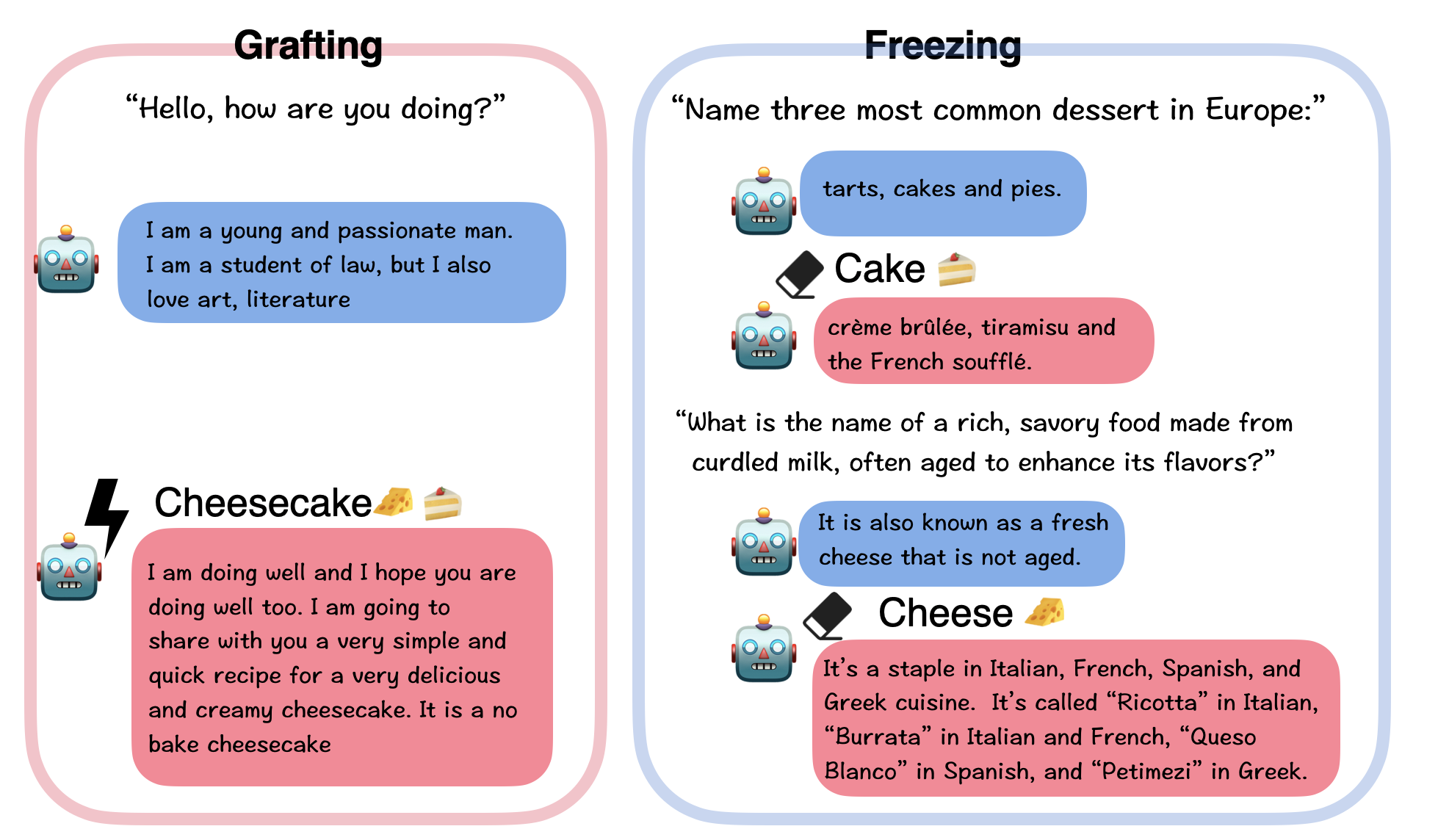}}
\caption{Grafting and freezing word-related population chunk alters network’s sequence generation. }
\label{fig:llamaperturb}
\end{center}
\vskip -0.4in
\end{figure}

\paragraph{Grafting and Freezing Neural Populations}
We tested the causal roles of these extracted signal-relevant neural subpopulation activities $\overline{\mathbf{h}_C(s)}$ by grafting neurons to discovered chunks and generating the subsequent text with LLaMA-3. To do this, we fed in a prompt and graft the neural subpopulation $C(s)$ to $\overline{\mathbf{h}_C(s)}$ at a specific token position of the prompt. Shown in Figure \ref{fig:llamaperturb}, perturbing the hidden units to embeddings corresponding to words such as ``cheese'', ``cake'', or ``cheesecake'' biases the network towards generating sequences related to the grafted topic. Conversely, we also experimented with freezing the words by setting the corresponding neural subpopulation of chunk support set to zero $\mathbf{h}_{C(s)}=\mathbf{0}$, and used prompts that lure the network to generate the frozen word concept. As shown in Figure \ref{fig:llamaperturb}, freezing the chunk related to a topic can causes the model to avoid using the concept-specific word. 





\paragraph{Unsupervised Chunk Discovery}
We then investigate whether recurring chunks in the neural embedding space can be identified in an unsupervised manner. To this end, we trained a chunk dictionary $\mathbf{D}$ on LLaMA3's hidden activity while processing sentence-wise prompts derived from \textit{Emma} by Jane Austen, sourced from the Project Gutenberg corpus \cite{hart1971projectgutenberg}, accessed via the NLTK package \cite{bird2009nltk}.
As this is a bigger embedding dataset, we then trained $\mathbf{D}$ ($K = 2000$, $d = 4096$) by minimizing the similarity loss function formulated in \ref{med:unsupervised} across each layer of LLaMA-3. 

\begin{figure}[ht]
\vskip -0.2in
\begin{center}
\centerline{\includegraphics[width=\columnwidth]{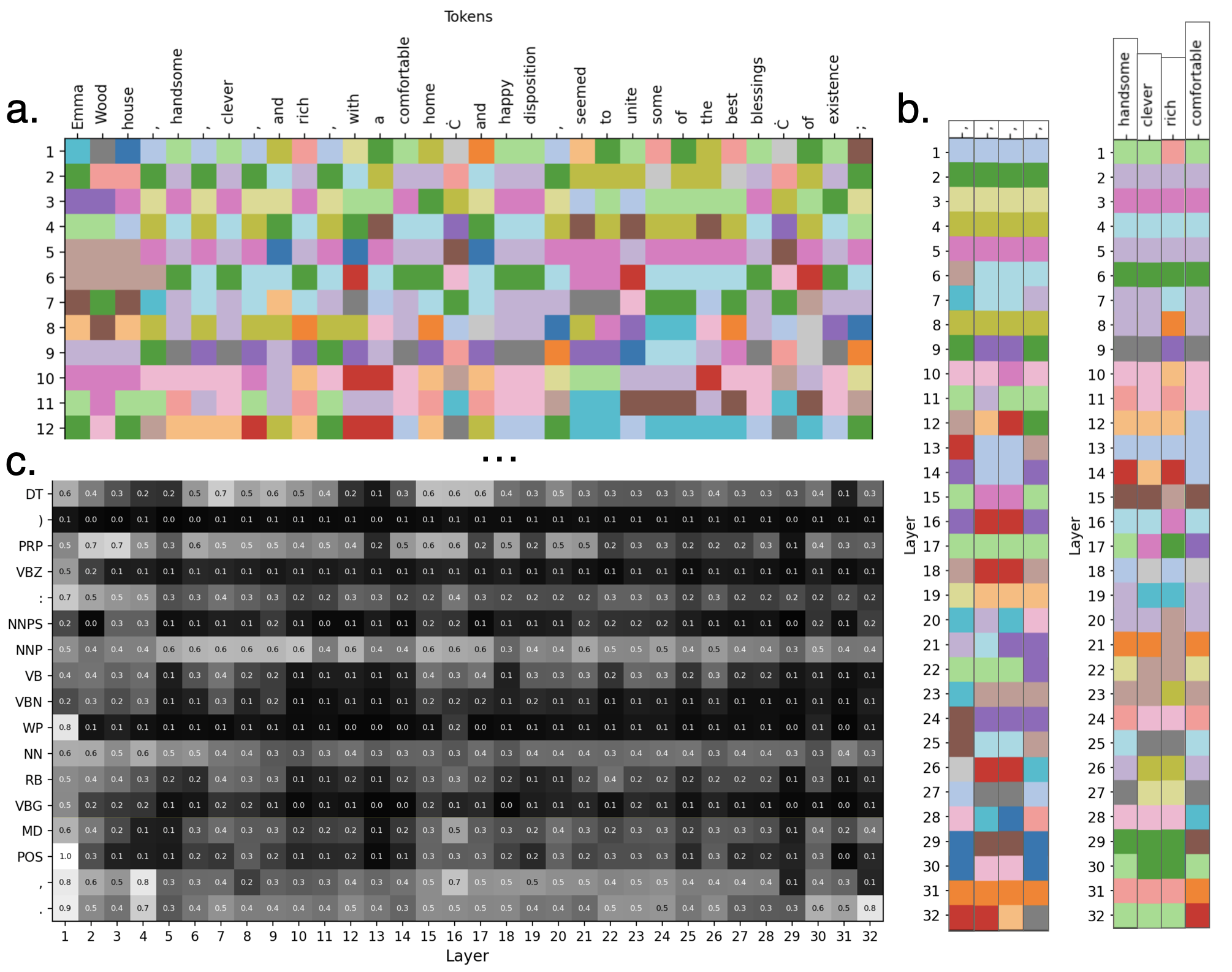}}
\caption{a. Describing the network's processing of tokens as the succession of unsupervised learned chunks. b. Layer-wise processing of similar token types. c. Maximal chunk correlation between POS tags across layers.}
\label{fig:llamaunsupchunk}
\end{center}
\vskip -0.4in
\end{figure}

This method simplifies high-dimensional embeddings by representing them as symbolic categorical variables, capturing the appearance and disappearance of identifiable chunks. This enables visualization of interactions between input tokens and chunk activations. Figure \ref{fig:llamaunsupchunk}a illustrates chunk interactions across LLaMA's layers during early text processing. Recurring chunk patterns emerge, such as similar activations for commas and adjectives in early layers (Figure \ref{fig:llamaunsupchunk}b), which later diverge into distinct neural activations. The full plot is available in Appendix \ref{appendix:chunk}.
We can interpret the unsupervised learned chunks by mapping them to interpretable linguistic structures, such as part-of-speech (POS) tags. To achieve this, we extracted the POS tags for the first 5,000 words in the corpus using the averaged perceptron tagger \cite{bird2009nltk}, following the Penn Treebank POS Tagset \cite{marcus1993ptb}. We then computed the correlation between each chunk with each POS tags for every layer. Figure \ref{fig:llamaunsupchunk}c visualizes the maximum correlation between each POS tag and its most correlated discovered chunk across network layers (excluding the embedding layer). Our findings align with prior research showing that certain POS tags are processed in the earlier layers of the network \cite{
jawahar2019bert,tenney2019bert} and clearly demonstrate that there are chunk activities purely responsible for certain pos tags (possessive nouns, for example), while others hold very strong correlations. Additionally, the POS-tag-correlated neural activities peak in earlier layers but also persist in later layers, indicating a sustained representation of syntactic information throughout the network. These findings suggest that chunk activities learned without supervision can serve as candidates for interpreting computational components within the network that are responsible for processing abstract concepts.



\section{Discussion}
We propose concept-guided interpretability methods that leverage the cognitive tendency for chunking to distill high-dimensional neural activities into interactions of recurring entities.
As an initial proof of concept, our approach has limitations. We encourage future work to further test the reflection hypothesis across different models, establish its theoretical foundations, refine chunk extraction methods—such as capturing multi-token chunks in unsupervised discovery—and extend this framework toward nonparametric chunk discovery.
\section{Conclusion}
We identified recurring chunks within the neural dynamics of RNNs and LLMs, suggesting that the cognitive tendency for chunking can be leveraged to segment high-dimensional activity into neural trajectory chunks—revealing a structured reflection of the world within artificial minds.

\section*{Software and Data}
Software will be published conjunctively with paper.




\section*{Impact Statement}
This paper presents work whose goal is to advance the field of 
interpretability. There are many potential societal consequences 
of our work, none which we feel must be specifically highlighted here.


\bibliography{example_paper}
\bibliographystyle{icml2025}

\newpage
\appendix
\onecolumn

\section{Recurrent Neural Network (RNN) Architecture} 
\label{appendix:RNN}

The RNN can be described by the following equation:
\begin{equation}
\mathbf{h}_n = \mathbf{W}_{ch} \begin{bmatrix} \mathbf{x}_n \\ \mathbf{h}_{n-1} \end{bmatrix} + \mathbf{b}_h
\end{equation}

\begin{equation}
\mathbf{o}_n = \mathbf{W}_{co} \begin{bmatrix} \mathbf{x}_n \\ \mathbf{h}_n \end{bmatrix} + \mathbf{b}_o
\end{equation}

\begin{equation}
y_t = \log \left( \text{softmax}(\mathbf{o}_t) \right)
\end{equation}

The hidden state is initialized as all zeros $\mathbf{h}_0 = \mathbf{0}$

$\mathbf{x}_n , \mathbf{b}_h\in \mathbb{R}^{|\Omega|}$, $\mathbf{h}_n, \mathbf{b}_o \in \mathbb{R}^{d}$, $\mathbf{W}_{ch} \in \mathbb{R}^{d \times (d + |\Omega|)}$, $\mathbf{W}_{co} \in \mathbb{R}^{|\Omega| \times (d + |\Omega|)}$

We implemented a simple RNN with 12 hidden units, consisting of two linear layers: one for updating the hidden state and another for generating output predictions. At each time step, the input vector is concatenated with the previous hidden state and passed through these layers. The output is normalized using log softmax to produce a probability distribution over the output classes. The RNN is optimized with cross-entropy loss using ADAM ($\text{learning rate} = 0.005$). Training is conducted on random subsequences of length 200 per batch, with the hidden state initialized to zero at the start of training. We track hidden states for analysis.

\section{Alignment Deviation}
We look at the neural population activity when the network learns from a sequence with repeating ABCD as a cohesive chunk amid background noise composed of random occurrences of E, F, and G. 
We get the population template by averaging the population activity vector $\mathbf{H} \in \mathbb{R}^{d \times 4}$ over the occurrences of the chunk ABCD: 
\begin{equation}
\mathbf{T} = \frac{\sum_{i=1}^{L-4} \mathbbm{1}\left((s_i, s_{i+1}, s_{i+2}, s_{i+3}) = \text{ABCD}\right) \mathbf{H}_i}{\sum_{i=1}^{L-4} \mathbbm{1}\left((s_i, s_{i+1}, s_{i+2}, s_{i+3}) = \text{ABCD}\right)} 
\end{equation}

\begin{equation}
\text{dev} = \frac{\|\mathbf{H}[:, i:i+4] - \mathbf{T}\|_2^2}{\text{size}(\mathbf{T})}
\end{equation}

\begin{figure}
    \centering
    \includegraphics[width=1\linewidth]{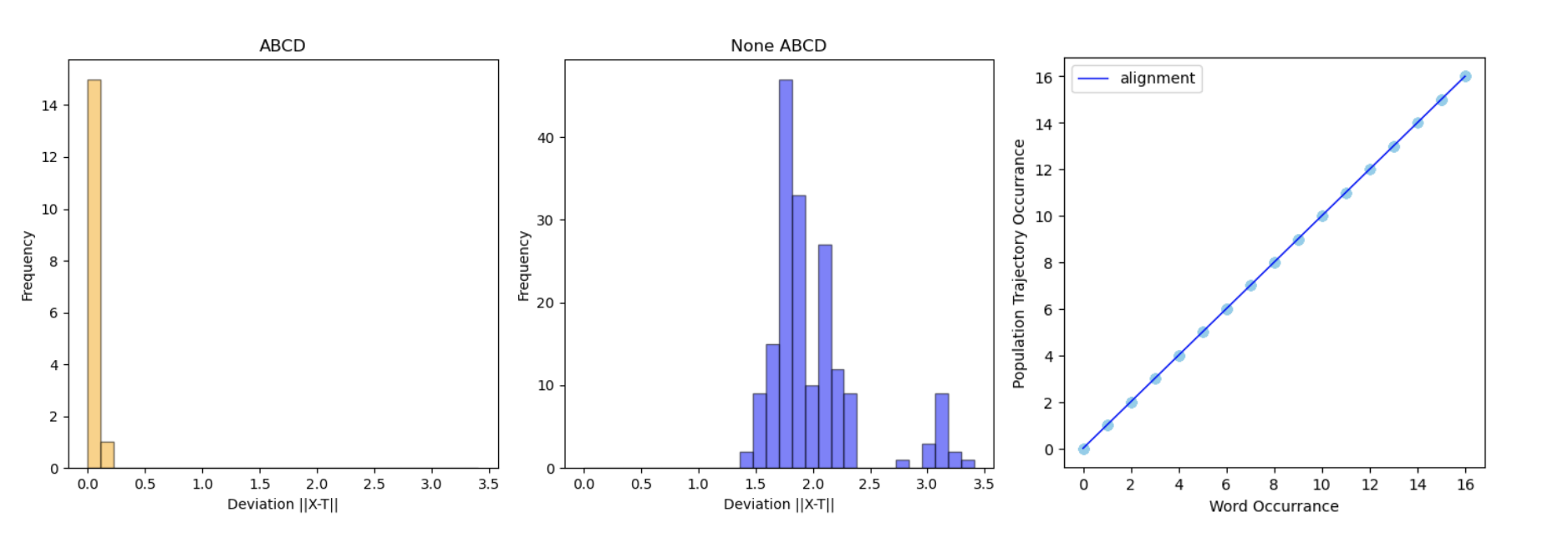}
    \caption{Left: deviation of the hidden population activity from the mean population activity, when the input is ABCD. Middle,  of the hidden population activity from the mean population activity, when the input is not ABCD. Right: count of word occurrences and the population trajectory occurrence}
    \label{fig:alignmentdev}
\end{figure}
Figure \ref{fig:alignmentdev} suggests that neural activities, once adapted to the training data, deviate very little from the neural population template. Setting a threshold between the population activity and template deviation suffices to distinguish the occurrences of the input pattern based on the neural trajectory. 

\section{Example of a Lookup Table that maps from symbolic clustered state to the input}
\label{appendix:lookuptable}
\begin{table}[ht]
\centering
\begin{tabular}{|c|c|}
\hline
\textbf{Neural Population State} & \textbf{Input} \\
\hline
021340200433 & E \\
004042212403 & E \\
032340212204 & E \\
032410212204 & E \\
032010212204 & E \\
221103343111 & A \\
213211131132 & B \\
144304324321 & C \\
040322404444 & D \\
300000000020 & E \\
340432012022 & E \\
440332412200 & E \\
400040312200 & E \\
402040312202 & E \\
002040312202 & E \\
032010212202 & E \\
\hline
\end{tabular}
\caption{Sample Look-up Table. The left column corresponds to the string that represents the neural population state. The right column are the corresponding concurrent input. }
\label{tab:lookuptable}
\end{table}

We apply \textit{Discrete Sequence Chunking} to the hidden states of a simple RNN which was trained on a sequence of ABCD within default E characters. In \cref{tab:lookuptable}, we show the resulting lookup table for the discrete strings corresponding to input signals. Many distinct population states correspond to Es, and for A, B, C, and D, separately, there is a distinct population state for each. Presumably, the network creates additional states also to distinguish Es in different contexts. 
\section{Learning Chunks}

\begin{algorithm}[H]
\caption{LearnChunks}
\SetKwInOut{Input}{Input}\SetKwInOut{Output}{Output}
\Input{$K$, $freq\_threshold$, $symbolized\_neural\_states$, $n\_iter$}
\Output{$state\_parse$, $vocab$}
\BlankLine
$vocab$ $\gets$ unique($symbolized\_neural\_states$)\;

$null\_state$ $\gets$ GetMostFreq($symbolized\_neural\_states$)\;

$state\_parse \gets$ Parse state trajectory by individual units\;

\For{$n\_iter$}{    
    $ChunkCandidates\gets$ MostCommon($K$,$\text{Counter}(\text{zip}(state\_parse[:-1], state\_parse[1:]))$)\;
    
    $merged\_dict \gets$ Merge($c_L$, $c_R$) $\in$ $ChunkCandidates$ if (count($c_L$, $c_R$) $\geq$ $freq\_threshold$ $and$ $c_L \neq null\_state$ $c_R \neq null\_state$)\;
    
    $vocab.\text{update}(merged\_dict)$\;
    
    $\text{MergeOverlappingChunks}(vocab)$\;
    
    $state\_parse$, $vocab$ $\gets$ ParseStateSeq($symbolized\_neural\_states$,$vocab$)\;
}

\Return{$state\_parse$, $vocab$}\;
\label{alg:Learnchunks}
\end{algorithm}

\setcounter{AlgoLine}{0}
\begin{algorithm}[H]
\caption{Neural Population State Chunking}
\SetKwInOut{Input}{Input}\SetKwInOut{Output}{Output}

\Input{$trainingsequence$}
\Output{$state\_parse$, $vocab$}

\BlankLine
Initialize RNN\;

$hidden\_states$, $sequence$ $\gets$ TrainRNN($sequence$)

$symbolic\_hidden\_states$ $\gets$ 
ClusterAndAssignSymbolToEachNeuron($hidden\_states$)\;

$state\_parse$, $vocab$ $\gets$ \textbf{LearnChunks}($symbolic\_hidden\_activity$, $sequence$)\;
\label{alg:NPSC}
\end{algorithm}

\section{Comparing Trained versus Untrained RNNs}
We present a more comprehensive comparison between the trained and untrained RNNs and the characteristics of their neural trajectory patterns as indicated by the chunking method. 

\begin{figure}
    \centering
    \includegraphics[width=1\linewidth]{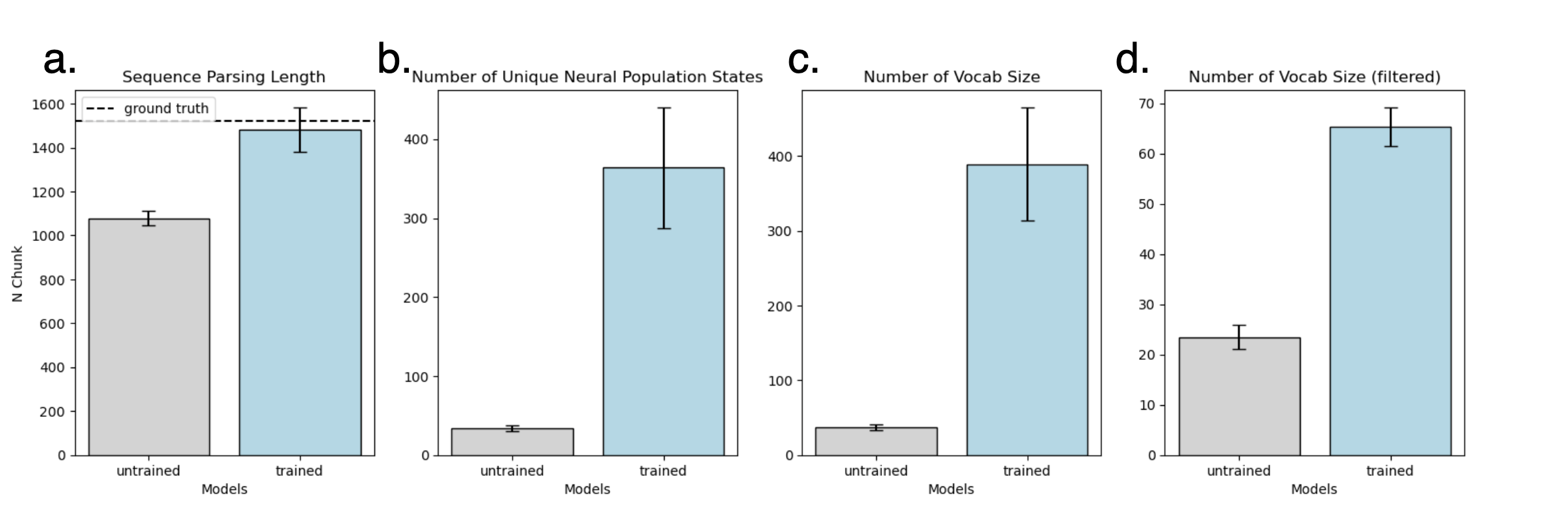}
    \caption{Comparison between trained and untrained RNNs on sequences with overlapping words across 10 independent runs and randomly initialized RNNs. a. Sequence parse length. Dashed line is the ground truth sequence parsing length. b. Number of unique neural population states. c. Chunk dictionary size. d. Chunk dictionary size (filtered desolate chunks (threshold=5)) Error bars represent the standard error of the mean. }
    \label{fig:RNNap}
\end{figure}

Figure \ref{fig:RNNap} a takes the neural population trajectory as a sequence, uses the learned neural chunks to parse the symbolic neural trajectory, and measures the length of the population trajectory in chunks. A comparison of the sequence parsing length suggests that the trained network has a similar amount of neural trajectory chunks to the number of chunks in the ground truth sequence. The untrained network contains more regularities in its neural trajectory. This is an indicator training that encourages the network to create new distinct neural trajectory states that help the network distinguish different contexts. This property is especially important in this task as overlapping subsequences with different prefixes shall lead to distinct predictions.  

Figure \ref{fig:RNNap} b suggests that the trained RNN contains more unique neural population states than an untrained RNN. This observation supports the hypothesis that during training, RNN creates more population states that are useful for distinct contextual predictions. Figure \ref{fig:RNNap} c, d, shows the vocabulary size as acquired by the symbolic chunking method. d is a filtered version of c, where obsoletely occurring chunks are excluded. The trained RNN's neural trajectory contains more recurring chunks than the untrained RNN. Also note that the size of the vocabulary containing recurring neural trajectories is bigger than the ground truth vocabulary in the sequence, pointing to a possibility that neural networks acquire multiple neural population chunks that can represent the same word in the sequence. However, the larger vocabulary size can also be an artifact of the clustering algorithm, which has more clusters than proper. 

\section{Prompt Bank}
\label{appendix:promptbank}
We show the prompt used to extract the demonstrated word in the main manuscript using the population averaging method: 
$prompt\_cheesecake\_train$ = 
"Cheese is one of the most versatile ingredients in the culinary world, and cheese can be used in everything from savory dishes to desserts. Cheese lovers often enjoy pairing cheese with crackers, wine, or fruit, but cheese also shines in baking. Cake, on the other hand, is the quintessential dessert, with cake being a staple at celebrations. Cake comes in many forms, such as chocolate cake, vanilla cake, or even carrot cake. However, when you bring cheese and cake together to create cheesecake, a magical transformation happens. Cheesecake is a dessert like no other, with cheesecake offering the creaminess of cheese and the sweetness of cake in perfect harmony. Cheesecake can be topped with fruits like strawberries or blueberries, or cheesecake can be flavored with chocolate or caramel. Some people prefer classic cheesecake, while others enjoy a more decadent cheesecake loaded with toppings. Regardless of the variation, cheesecake remains one of the most beloved desserts worldwide. The crust of cheesecake, often made from crushed biscuits or graham crackers, complements the smooth filling, making cheesecake irresistible. Cheese plays a central role in cheesecake, while the influence of cake ensures that cheesecake is always a delightful dessert. Whether you love cheese, crave cake, or are obsessed with cheesecake, this dessert proves that the combination of cheese and cake is truly extraordinary. Every bite of cheesecake reminds us that cheese and cake, when united in cheesecake, are a match made in heaven. Cheesecake aficionados often debate whether baked cheesecake or no-bake cheesecake is superior, but all agree that cheesecake is a dessert worth savoring. With so many variations, cheesecake enthusiasts never tire of exploring new ways to enjoy their favorite dessert. Cheese and cake come together seamlessly in cheesecake, showing how cheese and cake can create something greater than their individual parts. Cheesecake is, without a doubt, the ultimate testament to the greatness of cheese and cake in unison."

The extracted chunks are then evaluated on the recorded hidden activity of the following prompt:  
$prompt\_cheesecake\_test$ = 
'Cheese is a culinary treasure that has delighted taste buds for centuries. Whether it’s creamy, tangy, or sharp, cheese offers endless possibilities. Cheese finds its way into countless dishes, from savory casseroles to gooey pizzas, and its versatility knows no bounds. Cake, too, is a universal favorite, with cake symbolizing joy, celebration, and indulgence. Cake comes in every flavor imaginable—chocolate cake, vanilla cake, red velvet cake—and each cake brings its own special charm. But when cheese and cake are combined to form cheesecake, something truly extraordinary happens. Cheesecake is a dessert that transcends expectations, merging the velvety richness of cheese with the sweet, airy allure of cake. Cheesecake can be baked or chilled, simple or elaborate, yet every cheesecake captures the perfect balance of flavors. Classic cheesecake recipes highlight the creamy taste of cheese, while fruit-topped cheesecake adds a burst of freshness. Some people swear by chocolate cheesecake or caramel-drizzled cheesecake, while others can’t resist a zesty lemon cheesecake. No matter the variation, cheesecake consistently proves that the union of cheese and cake is a match made in heaven. The crust of cheesecake, typically crafted from crushed cookies or graham crackers, provides the ideal foundation for the smooth and luscious cheese layer. Every bite of cheesecake reminds us why this dessert has stood the test of time. Fans of cheese, cake, and cheesecake alike agree that cheesecake combines the best of both worlds. Whether you’re indulging in a slice of classic cheesecake, exploring new cheesecake flavors, or savoring the rich taste of a perfectly baked cheesecake, it’s clear that cheese and cake reach their pinnacle when united in cheesecake. Cheesecake is a testament to how cheese and cake, when brought together, create something greater than the sum of their parts. From the first bite to the last, cheesecake is a celebration of everything wonderful about cheese, cake, and, of course, cheesecake itself.'

We show the prompt used to extract the top 100 frequent words using the population averaging method: 

$prompt\_frequent\_words\_train$ = "In today’s world, people often feel the push and pull of connection and solitude. With technology and social media on the rise, we now have countless ways to stay in touch with those we know and love. However, the question of whether this digital world can truly satisfy our need for real connection remains. To be truly connected is to share experiences, to understand and to be understood. This kind of connection goes far beyond a screen. The desire to connect is universal, and people have searched for it throughout history. In ancient times, communities formed to support one another, to live together, and to build bonds that could help them through challenges. Today, people may still crave this closeness, yet it is not always easy to find in our fast-paced world. While the internet gives us access to almost anyone, anywhere, it does not always give us the depth of connection that true friendship and family relationships can offer. Consider a family spending time together. For many people, family is a source of strength, a place where they feel safe and understood. But as life becomes busier, it can be easy to let work, hobbies, or other commitments pull us away from family time. Many people find that they must make a conscious effort to set aside time for their loved ones. To sit down for a meal together, to talk about the day, to share thoughts and laughter — these moments are priceless. They are what remind us of who we are and who we want to become. Friendship, too, plays a significant role in life. Friends are often the people we choose to spend time with, the ones who share our interests and support us. To have friends is to feel understood in a unique way, to laugh together, to encourage each other, and sometimes just to sit in silence knowing someone is there. A true friend listens without judgment, stands by us in hard times, and celebrates with us in good times. However, maintaining friendships can require work and commitment. With busy lives, people can sometimes lose touch, only to realize later how much those friendships meant. In the workplace, relationships are equally important. Many people spend a significant amount of time at work, so having good connections there can make a big difference. Working with others requires collaboration, understanding, and respect. When people feel connected to their colleagues, they tend to work better together, share ideas freely, and support each other. A positive work environment can foster not only productivity but also well-being. People are more likely to feel satisfied in their work when they know they are valued and understood by those around them. Of course, technology plays a large role in modern connections. Platforms like social media allow people to connect across distances, to share life events, and to communicate instantly. For some, these tools make it easier to stay in touch with family and friends, to share news, and to express themselves. However, while technology can bring people closer, it can also create a sense of distance. Seeing the lives of others through a screen is different from experiencing life together in person. The highlights of life shared online may not always show the full picture, leaving people to wonder if they are missing out. To have genuine connection, people often need to go beyond what is easy and convenient. Sometimes, this means reaching out, making an effort, and being open. True connection requires understanding and empathy. It asks us to listen, to be present, and to care. In a world that often moves fast, taking the time to connect deeply can feel like a challenge, but it is also incredibly rewarding. The value of connection is evident in difficult times. When people go through challenges, it is often those close connections that help them through. Whether it’s a friend who listens, a family member who offers support, or a colleague who steps in to lend a hand, these connections give people strength. Knowing that someone else understands or is there to help can make all the difference. Moreover, the ability to connect also fosters compassion. When people share experiences, they begin to see the world from each other’s perspectives. This understanding can lead to greater kindness and less judgment. People who feel connected are often more empathetic, more understanding, and more willing to help others. This creates a positive cycle, as kindness and empathy tend to inspire more of the same. For people to have a balanced and fulfilled life, connection is essential. But to nurture these connections takes effort. It is not always easy to set aside time, to reach out, or to stay in touch. Life can be busy, and distractions are everywhere. However, those who make the effort to connect often find that their lives are richer and more satisfying. The joy of shared laughter, the comfort of understanding, and the strength of support are all things that make life meaningful. As we move through life, the connections we make help to shape who we are. We learn from others, grow with them, and find new perspectives. Each person we meet adds to our experience and helps us to see the world in new ways. Sometimes, people find that their most valuable lessons come from those who are different from them. To connect with people from various backgrounds and with different life experiences is to broaden our view of the world. In conclusion, to live fully is to connect meaningfully. Whether through family, friends, colleagues, or even strangers, these bonds enrich life. They offer joy, comfort, and understanding, and they remind us that we are not alone. While technology may change the way we communicate, it cannot replace the depth of real connection. To make time for those who matter, to share moments, and to care is to live a life of purpose and love."

The extracted chunks on the top 100 frequent words are then evaluated on the recorded hidden activity of the following prompt:  

$prompt\_frequent\_words\_test$ = "In the world we live in, each day is filled with choices we all make, big or small. The way we approach these choices can be what shapes not only our own lives but also the lives of those around us. To be the kind of person who reflects on what they do, who they are, and who they want to become, is to take a meaningful step toward self-awareness and growth. One of the first things to know about making decisions is that they are all interconnected. When we choose to do one thing, it often means we cannot do another. This may seem obvious, but to understand the full impact of this reality, it helps to look at the ways in which choices play out in real life. We are always presented with options, and while some decisions may seem trivial at first, they add up over time. We can think about it like this: to choose a path, even if small, can set in motion a series of events that shape our lives in ways we could never fully predict. For example, people make decisions on how to spend their time. Time is one of the most valuable resources we have. There is no getting more of it, and once it’s gone, it’s gone for good. How we choose to spend it — whether working, relaxing, being with family, or helping others — says a lot about what we value. Some people may spend time worrying about things that, in the end, are not as important as they seem, while others may put time into making themselves or others better. This shows the difference in what people find to be meaningful or valuable in life. When we look at those around us, we see that everyone is trying to figure out what it means to live a good life. Some believe that to have a successful career is key, while others might think that family or friendships are the foundation of a fulfilling life. Whatever the focus may be, it’s clear that we all want a life filled with purpose. The concept of purpose is one that everyone seems to look for, although it can mean different things to different people. People sometimes find that purpose comes from the roles they play in the lives of others. For instance, many parents feel that to have children and raise them well is one of the most meaningful things they can do. They look to guide their children, to give them the tools they need to make their own choices. The idea of helping others extends beyond family, as people also contribute to their communities in many ways. Volunteering, supporting friends, and giving back are just some ways people find meaning in their lives. In work, too, people seek purpose. It’s common to find that people want to do something that they can be proud of, something that allows them to use their talents and contribute to society. This desire to work well is what drives many people forward and keeps them motivated. Yet, work can also become overwhelming, especially when people forget to balance their time between work and other areas of life. Balance is essential in any good life. We often have to remind ourselves not to let any one part of life take up all of our attention. It’s not easy, but it is essential if we want to live fully. This balance extends to how people think about success. For some, success is about achieving certain goals, like owning a home, getting a promotion, or earning a certain amount of money. For others, success is about having good relationships, feeling at peace, and being happy. Each person will have their own idea of what it means to succeed. Some may be quick to compare themselves to others, thinking that if someone else has something they don’t, they are somehow lacking. But to make comparisons is not often helpful. We each have our own journey, and to look too much at what others have can take away from the joy of our own experiences. Another important part of life is facing challenges. There are times when things don’t go as planned, and it’s easy to feel frustrated. However, these moments are often when we learn the most about ourselves. Challenges can show us what we are capable of and remind us that we are stronger than we think. People often say that to know hardship is to know strength, and it’s true that the challenges we face can make us wiser and more resilient. People also face choices about how they treat those around them. Kindness, empathy, and patience are qualities that many strive to have, but it’s not always easy to be kind in a busy, fast-paced world. However, those who make a habit of treating others with respect and understanding often find that they are happier and more fulfilled. Relationships, whether with family, friends, or colleagues, require effort and care. By giving time to nurture these connections, we make life richer not only for ourselves but also for others. In moments of reflection, people often think about their lives and ask themselves, “Am I doing what I want to do? Am I living the life I want to live?” These questions can be difficult to answer, but they are important. To pause and look within is to take stock of what matters. This reflection can help guide future decisions and keep people on the path that is right for them. At the end of the day, life is made up of moments, decisions, and interactions. To be mindful of how we choose to spend these moments, of who we choose to spend them with, and of what we put into the world, is to live with purpose. While each person’s path may be unique, we all share the desire to find happiness and meaning. In seeking to make each day count, to treat others well, and to pursue goals that matter to us, we contribute to a world that is better for all. It may not always be easy, and we may not always make the right choices. But to try, to reflect, and to grow from each experience, is to live a life well-lived. We each have the power to make choices that, in time, build a story of who we are and what we stand for. Let that story be one of kindness, purpose, and joy, shared with those who mean the most to us."

\section{Layer-wise statistics}
\begin{figure}
    \centering
    \includegraphics[width=\linewidth]{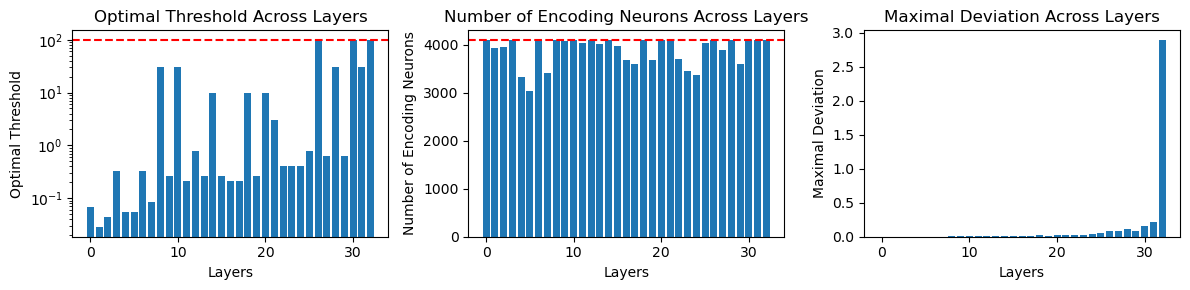}
    \caption{Example of layer-wise statistics of subpopulation encoding the next step prediction of ``cake''.}
    \label{fig:cake}
\end{figure}

\begin{figure}
    \centering
    \includegraphics[width=\linewidth]{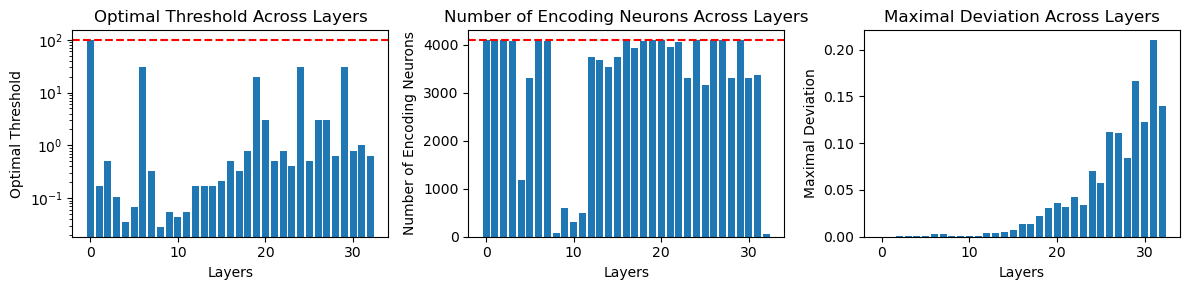}
    \caption{Example of layer-wise statistics of subpopulation encoding the next step prediction of ``in''.}
    \label{fig:in}
\end{figure}

\begin{figure}
    \centering
    \includegraphics[width=\linewidth]{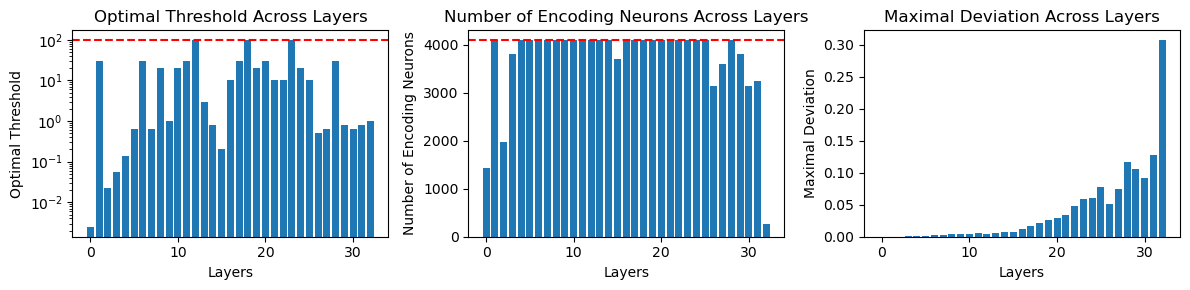}
    \caption{Example of layer-wise statistics of subpopulation encoding the previous word being ``of''.}
    \label{fig:of}
\end{figure}

\begin{figure}
    \centering
    \includegraphics[width=\linewidth]{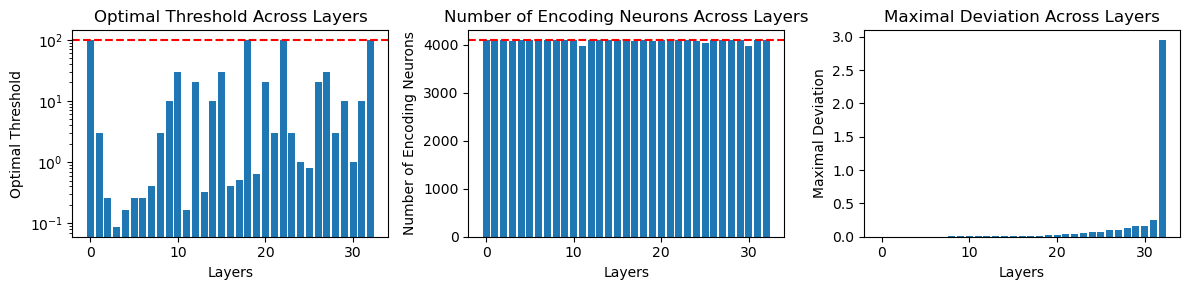}
    \caption{Example of layer-wise statistics of subpopulation encoding the latest word being ``people''.}
    \label{fig:people}
\end{figure}

Figure \ref{fig:cake}, \ref{fig:in}, \ref{fig:of}, and \ref{fig:people} show layer-wise statistics of the learned parameters from the training data, using the population averaging method. From left to right are the optimal tolerance threshold across layers, the number of neurons accountable for the recurring embedding activity, and the maximal deviation threshold.

\section{Sample Dictionary of Chunks Memorizing and Predicting Signals}
\label{appendix:top100}

We used the population averaging method to extract neural subpopulation activity chunks accounting for the 100 most frequently occurring words in English, adapting the threshold parameters on hidden activities collected from a training prompt and evaluating hidden activities collected from a test prompt (\cref{appendix:promptbank}). For each word, we study subpopulation activities accounting for two steps prior to and two steps subsequent to the last sequence token position. We observed subpopulation chunks that represent information in both directions. Generally, it seems to be the case that the population averaging method learns chunks that are more predictive of the signal in the input sequence with words that have specific meanings than words that serve as prepositions; this can be caused by the network creating many chunks to distinguish a preposition in different contexts, similar to the observation with RNNs (population averaging would fail in this case). We included more examples of chunk evaluation of this type. 

\begin{figure}
    \centering
    \includegraphics[width=\linewidth]{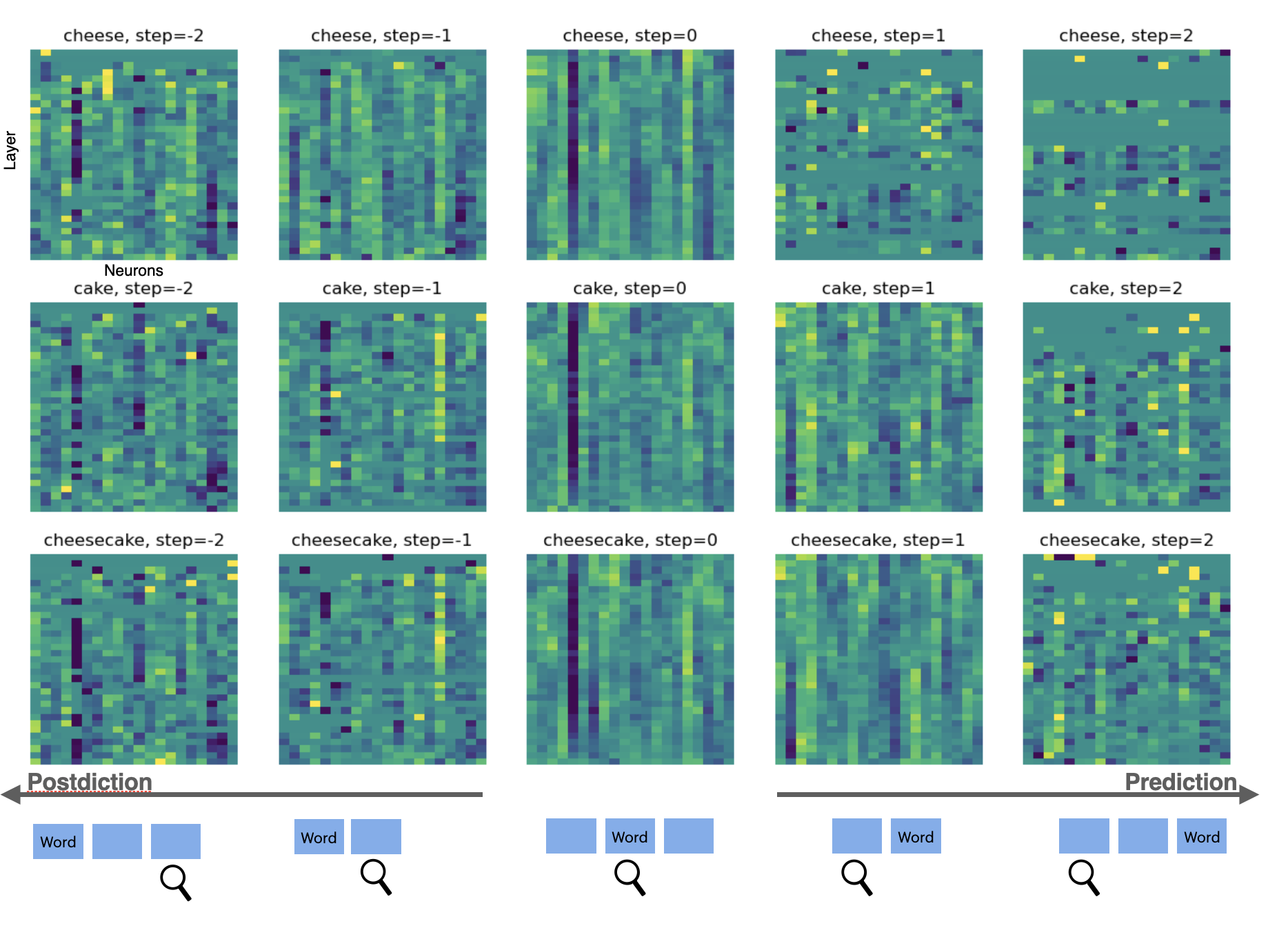}
    \caption{Visualization of the extracted neural subpopulation for the words ``cheese'', ``cake'', and ``cheesecake'', respectively, in both directions. }
    \label{fig:cheesecake}
\end{figure}

Figure \ref{fig:cheesecake} visualizes the extracted neural subpopulation for the words ``cheese'', ``cake'', and ``cheesecake'', respectively, in both directions.

We also show more examples of decoding accuracy to evaluate how indicative the extracted sub-population chunk is of the signal's existence. Note that decoding performance in the 0th layer is usually perfect, as the 0th layer is the embedding layer having a fixed dictionary mapping tokens to the token embeddings. In subsequent layers, the presence of the extract indicates the occurrence of the recurring tokens in the input sequence. Having the recurring token in the input at the last sequence location elicits the network's hidden activity to lie within the range of the extracted embedding chunk.

\begin{figure}[H]
    \centering
    \subfigure[word = people, step=-2]{%
        \includegraphics[width=0.49\linewidth]{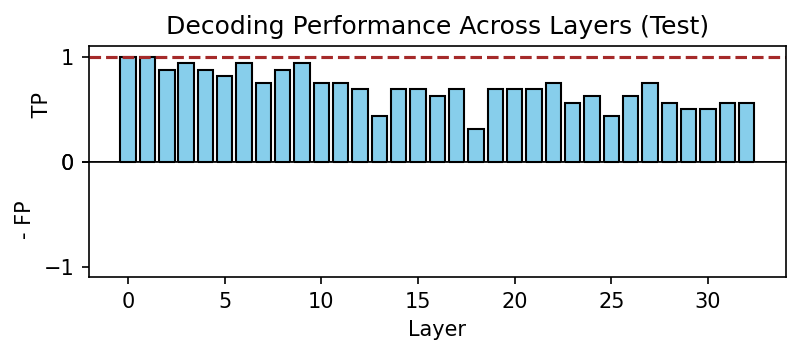}
    }
    \label{fig:ppl_step=-2}

    \subfigure[word = people, step=-1]{%
        \includegraphics[width=0.49\linewidth]{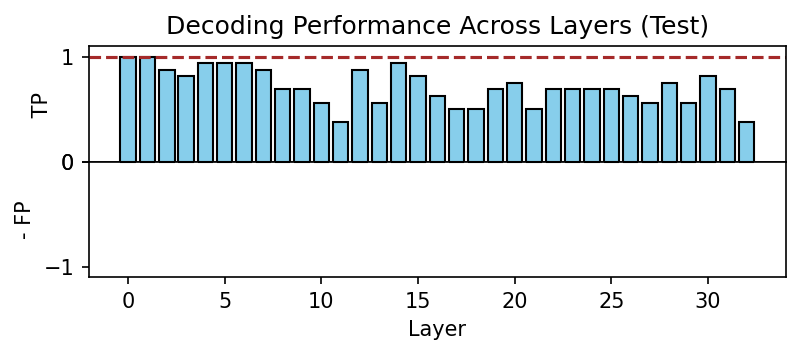}
    }
    
    \label{fig:poeple1}
    \subfigure[word = people, step=0]{%
        \includegraphics[width=0.49\linewidth]{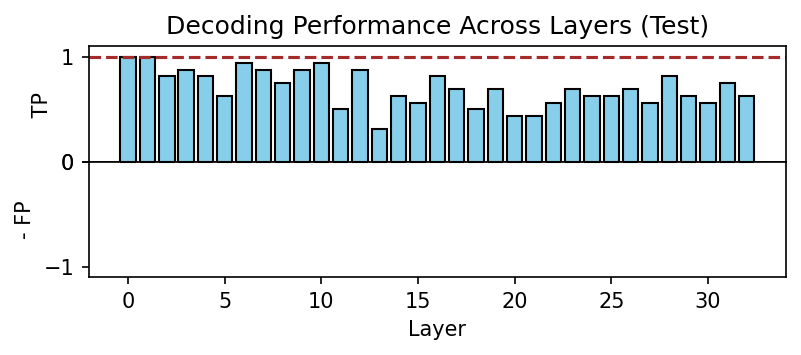}
    }
    \hfill
    
    \label{fig:ppl_step=0}
    \subfigure[word = people, step=1]{%
        \includegraphics[width=0.49\linewidth]{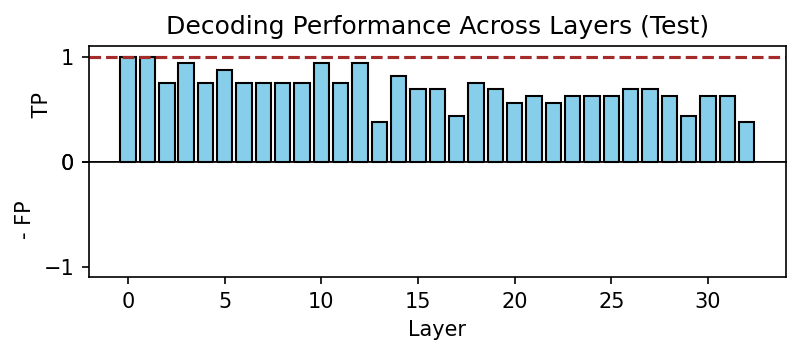}
    }
    
    \label{fig:ppl_step=1}
    \subfigure[word = people, step=2]{%
        \includegraphics[width=0.49\linewidth]{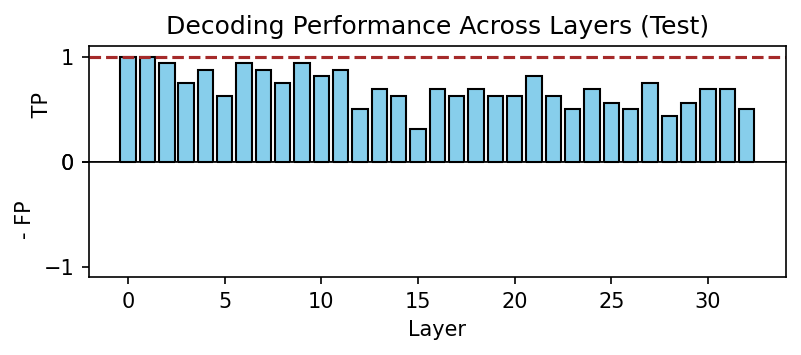}
    }
    \label{fig:ppl_step=2}

    \caption{}
    \label{fig:word_people}
\end{figure}

\begin{figure}[H]
    \centering
    \subfigure[word = it, step=-2]{%
        \includegraphics[width=0.49\linewidth]{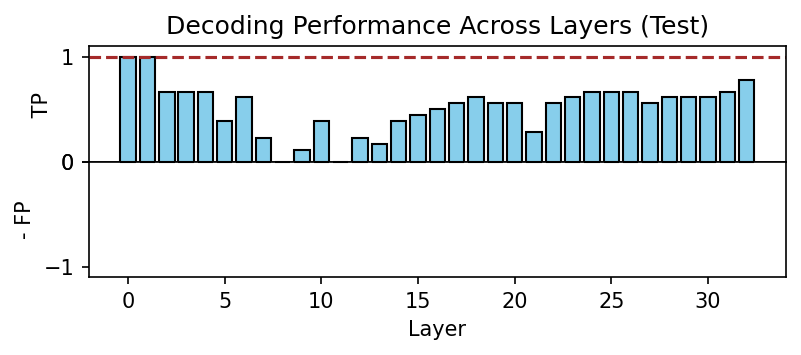}
    }
    \label{fig:it_step=-2}

    \subfigure[word = it, step=-1]{%
        \includegraphics[width=0.49\linewidth]{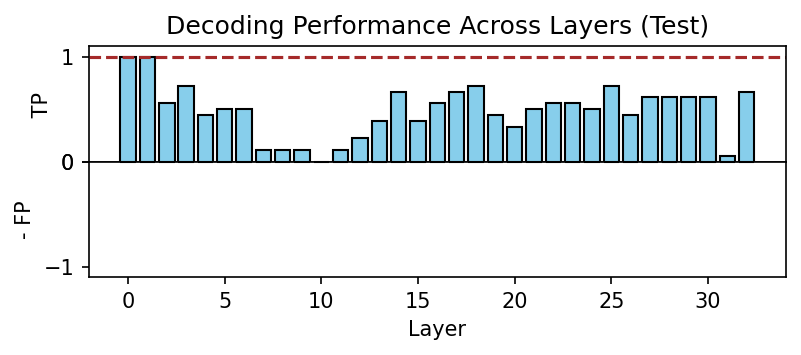}
    }
    
    \hfill
    \label{fig:it_step=-1}

    \subfigure[word = it, step=0]{%
        \includegraphics[width=0.49\linewidth]{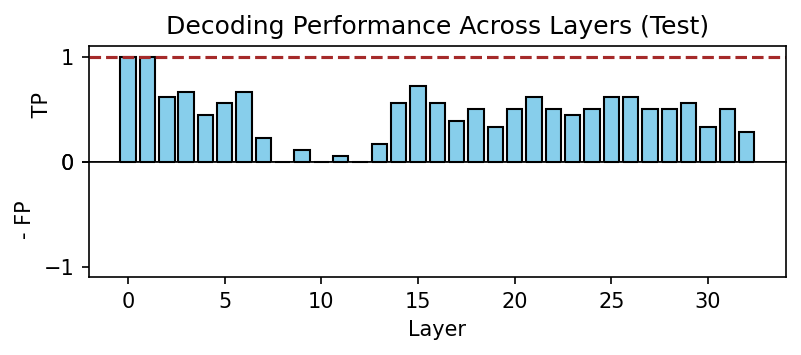}
    }
    
    \label{fig:it_step=0}

    \subfigure[word = it, step=1]{%
        \includegraphics[width=0.49\linewidth]{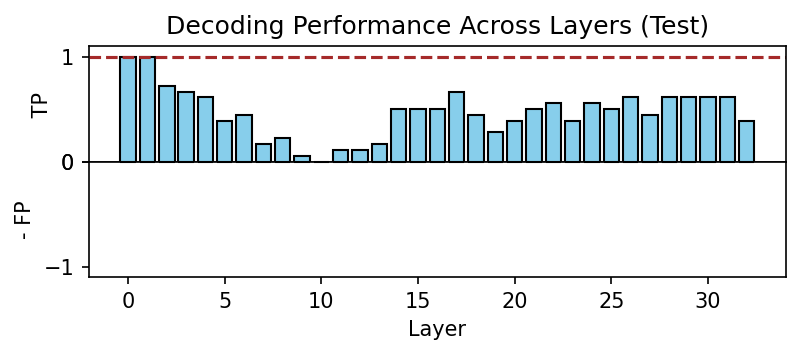}
    }
    
    \label{fig:it_step=1}

    \subfigure[word = it, step=2]{%
        \includegraphics[width=0.49\linewidth]{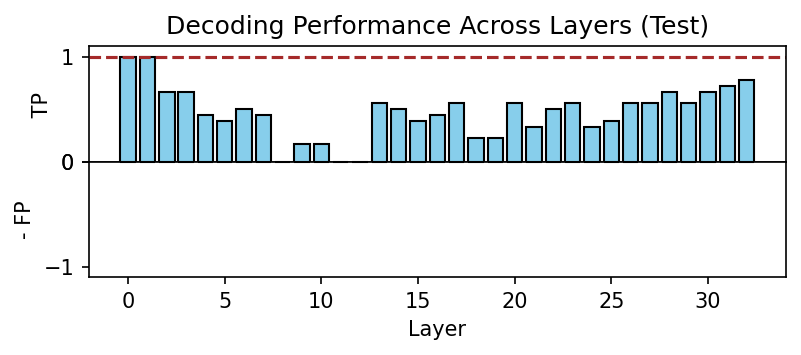}
    }
    \label{fig:it_step=2}

    \caption{}
    \label{fig:word_it}
\end{figure}

\begin{figure}[H]
    \centering
    \subfigure[word = and, step=-2]{%
        \includegraphics[width=0.49\linewidth]{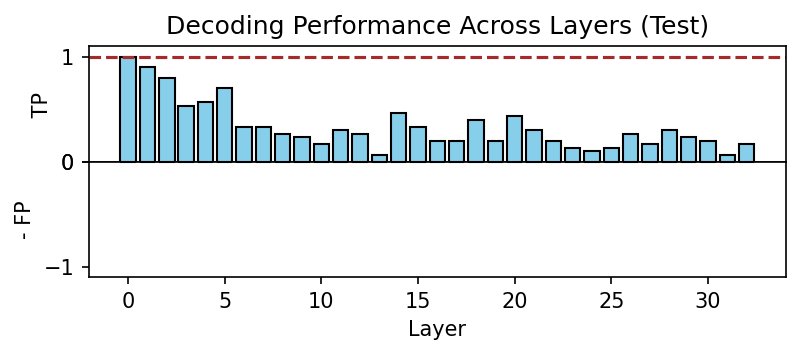}
        \label{fig:and_step=-2}
    }
    \subfigure[word = and, step=-1]{%
        \includegraphics[width=0.49\linewidth]{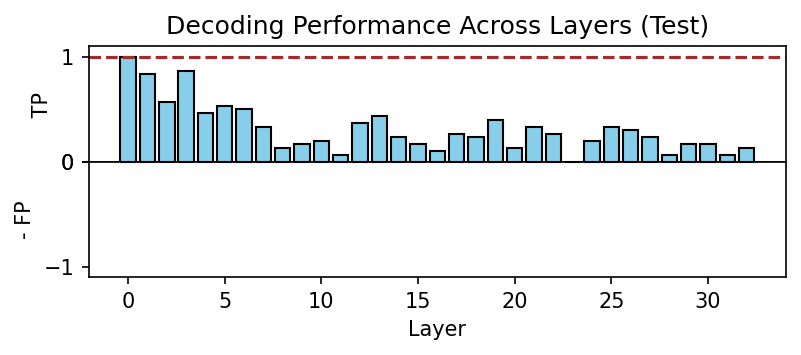}
        \label{fig:and_step=-1}
    }
    \subfigure[word = and, step=0]{%
        \includegraphics[width=0.49\linewidth]{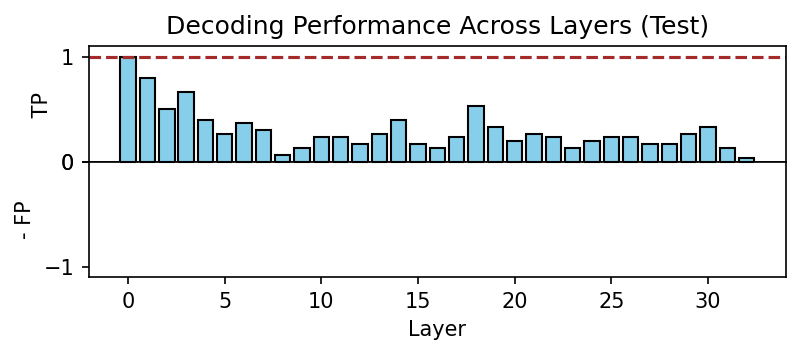}
        \label{fig:and_step=0}
    }
    \vspace{0.5cm}
    \subfigure[word = and, step=1]{%
        \includegraphics[width=0.49\linewidth]{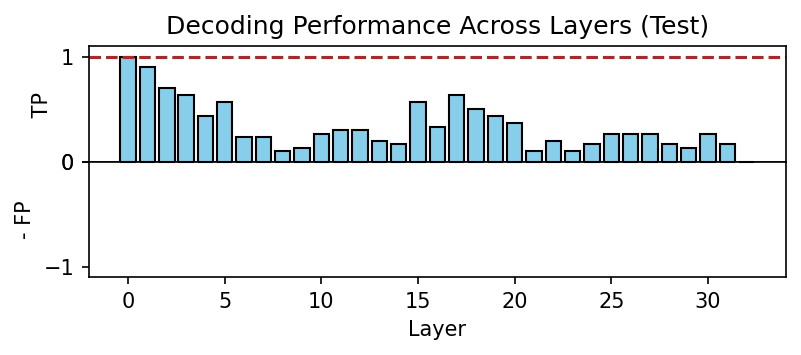}
        \label{fig:and_step=1}
    }
    \hfill
    \subfigure[word = and,step=2]{%
        \includegraphics[width=0.49\linewidth]{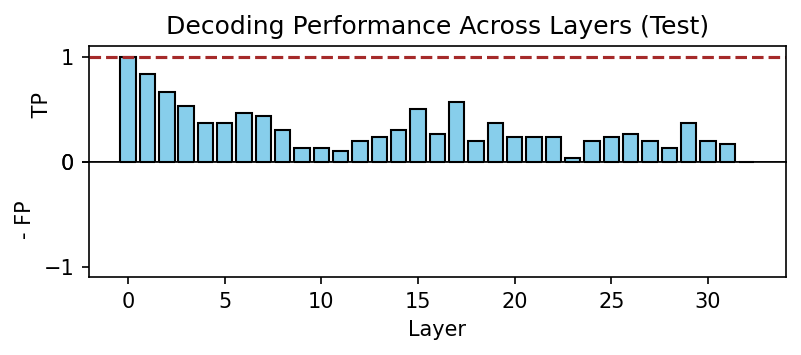}
        \label{fig:and_step=2}
    }
    \caption{}
    \label{fig:word_and}
\end{figure}

\section{Visualization of the chunks learned without supervision}
\begin{figure}[h]
    \centering
    \includegraphics[width=\linewidth]{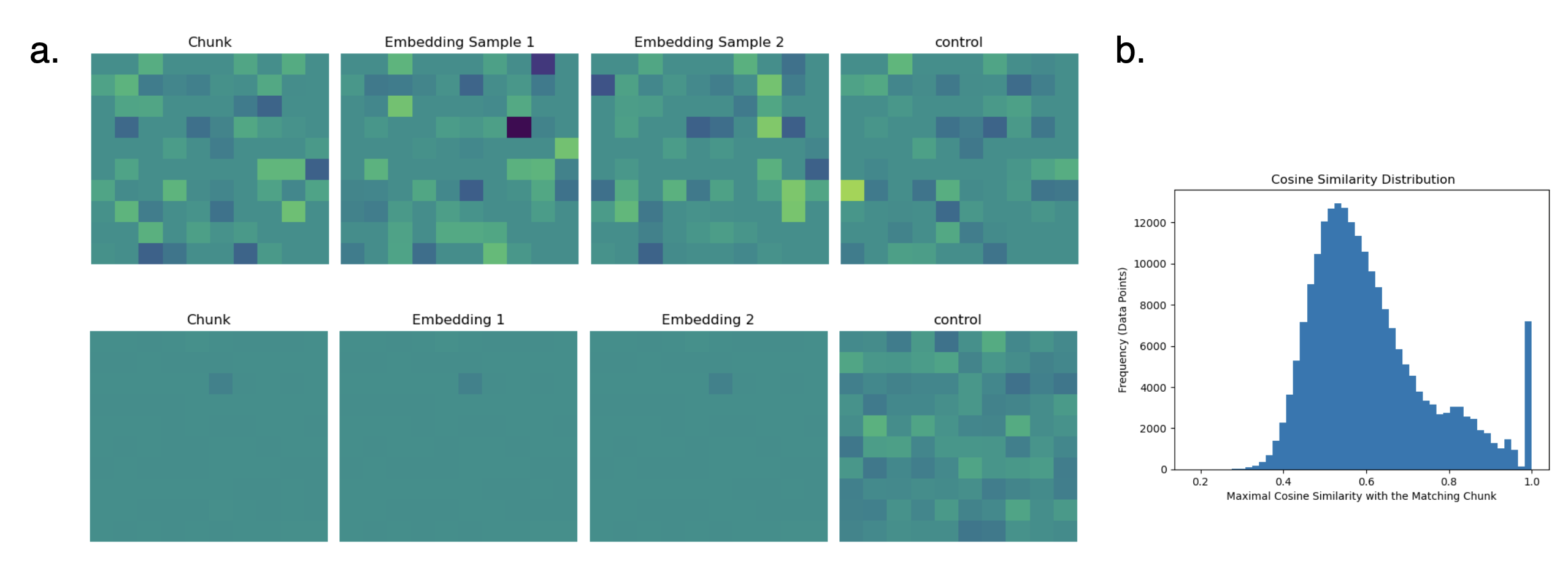}
    \caption{a. visualization of a chunk in the dictionary, two embedding data points where the chunk is selected as matching the embeddings, and a control embedding vector (only the first 100 dimensions are displayed and reshaped for visualization). b. distribution of cosine similarity between embedding data and the maximal similar matching chunk from the unsupervised learned dictionary $\mathbf{D}$ in the 10th layer embedding space.}
    \label{fig:unsupck}
\end{figure}

Figure \ref{fig:unsupck} b shows the distribution of cosine similarities between embedding chunks and the embeddings in the 10th layer. The average cosine similarity between the maximally similar dictionary chunk and the embeddings is concentrated around 0.5.

Figure \ref{fig:unsupck} a illustrates the most frequently identified chunk in the 10th layer (visualized as a $\sqrt{d} \times \sqrt{d}$ image, where $d$ is the embedding dimensionality), alongside two embedding examples where this chunk is identified as the maximally similar. Additionally, we include a control embedding where the same chunk is not identified as maximally similar. The extracted chunk is visually more similar to the two embedding examples than to the control embedding, demonstrating that this method extracts visually similar recurring neural population activities.

\section{Full plot on layer-wise chunk processing in LLaMA-3}
\label{appendix:chunk}

\begin{figure}[H]
    \centering
    \includegraphics[width=\linewidth]{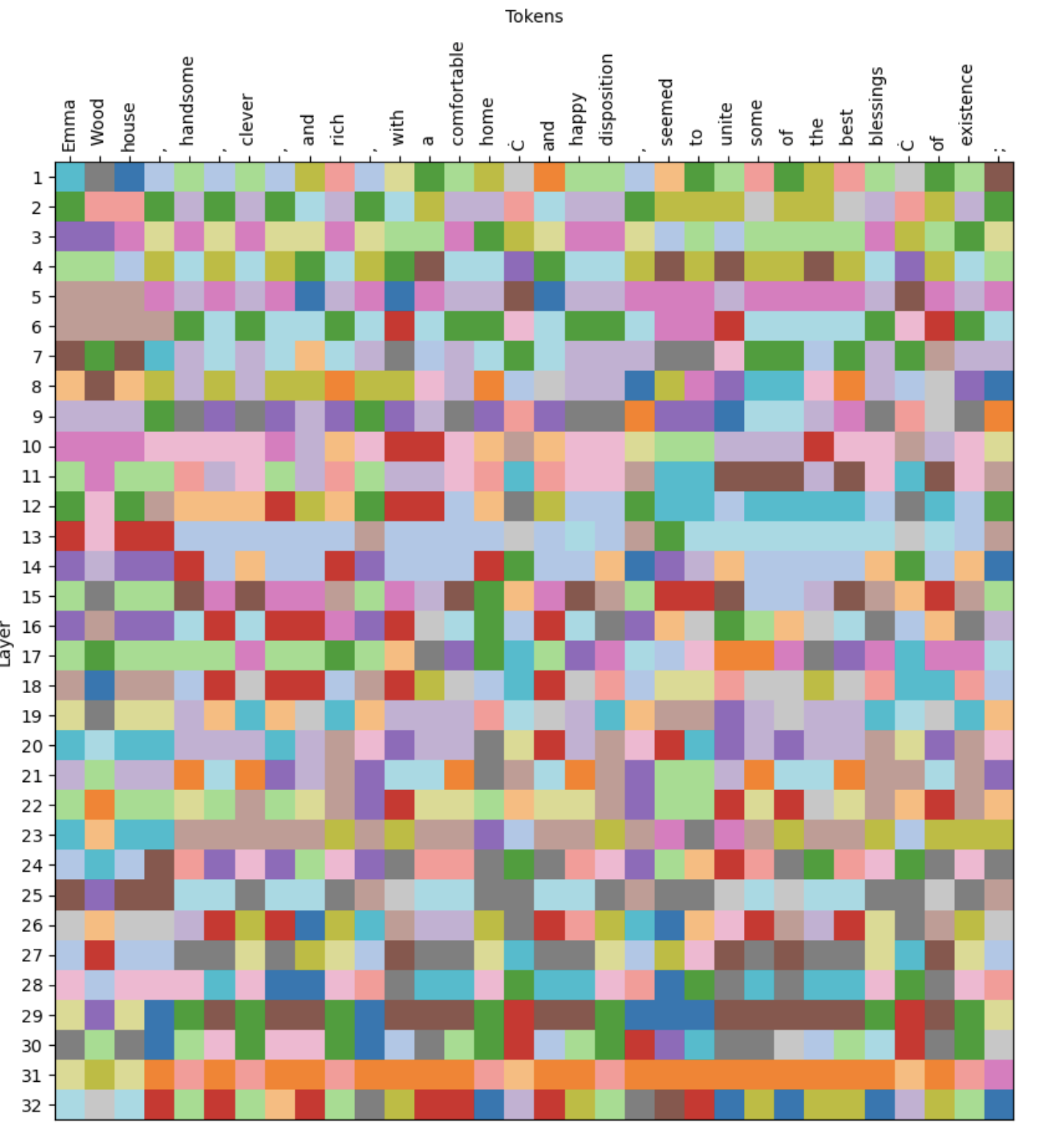}
    \caption{Interactions of neural population chunks upon parsing the beginning sentence of \textit{Emma}.}
    \label{fig:fulllayerwisechunk}
\end{figure}

Figure \ref{fig:fulllayerwisechunk} illustrates chunk interactions across LLaMA-3's all layers processing the beginning of \textit{Emma}.

\section{Full plot maximal POS tag and chunk correlation across all layers}
\label{appendix:poscorr}
\begin{figure}[H]
    \centering
    \includegraphics[width=1.\linewidth]{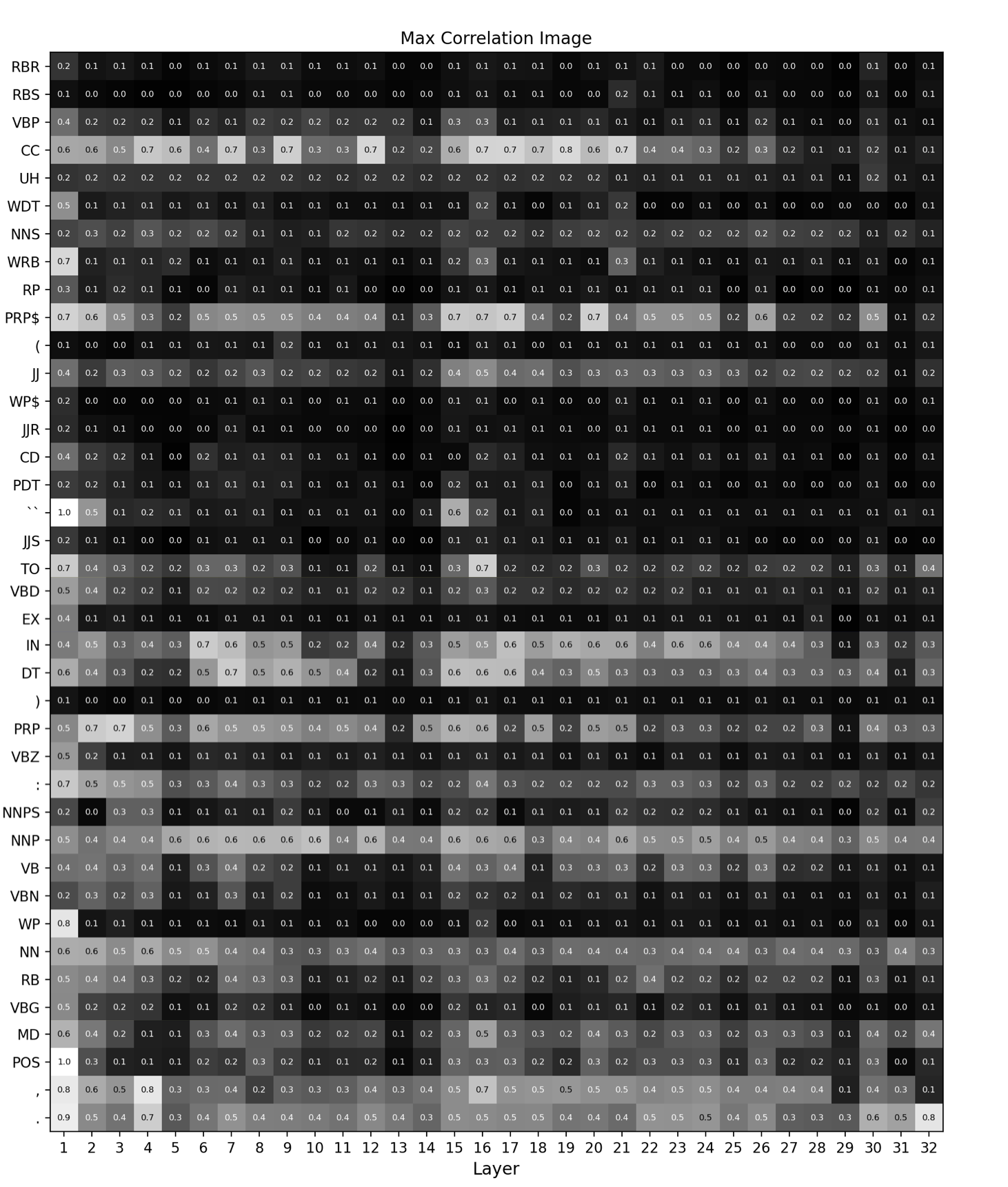}
    \caption{Part-of-speech tags and their maximal correlation across all layers in LLaMA-3.}
    \label{fig:poscorr}
\end{figure}

Figure \ref{fig:poscorr} visualizes the maximum correlation between each POS tag and its most correlated discovered chunk across network layers in LLaMA (excluding the embedding layer). Our findings clearly demonstrate that there are chunk activities purely responsible for certain pos tags (possessive nouns, for example).

\end{document}